\documentclass[3p,times]{article}
\usepackage{csl}

\usepackage{amssymb}

\usepackage[figuresright]{rotating}

\usepackage{physics}
\usepackage{dsfont}
\usepackage{amsmath}
\usepackage{amssymb}
\usepackage{amsthm}
\usepackage{amsfonts}
\usepackage{mathtools}
\usepackage{mathrsfs}
\usepackage{hyperref}
\usepackage{cleveref}
\usepackage{bm}
\usepackage{lipsum}
\usepackage{graphicx}
\usepackage{listings}
\usepackage{float}
\usepackage{wrapfig}
\usepackage{caption}
\usepackage{subcaption}
\usepackage{lineno}
\usepackage[square, numbers]{natbib}

\usepackage[table,dvipsnames]{xcolor}

\usepackage[listings,skins,breakable]{tcolorbox}

\theoremstyle{definition}
\newtheorem{definition}{Definition}

\theoremstyle{definition}
\newtheorem{remark}{Remark}

\newif\ifcomplete
\newif\ifdraft

\providecommand{\x}[1][]{%
   \ifthenelse{ \equal{#1}{} }
      {\mathbf{x}}
      {\mathbf{x}^{[#1]}}}
\providecommand{\xb}[1][]{%
   \ifthenelse{ \equal{#1}{} }
      {\mathbf{x}^{\rm b}}
      {\mathbf{x}^{{\rm b}[#1]}}}
\providecommand{\xa}[1][]{%
   \ifthenelse{ \equal{#1}{} }
      {\mathbf{x}^{\rm a}}
      {\mathbf{x}^{{\rm a}[#1]}}}
\providecommand{\xf}[1][]{%
   \ifthenelse{ \equal{#1}{} }
      {\mathbf{y}}
      {\mathbf{y}^{[#1]}}}
\providecommand{\y}[1][]{%
   \ifthenelse{ \equal{#1}{} }
      {\mathbf{y}}
      {\mathbf{y}^{[#1]}}}

\providecommand{\bigdot}[1]{\accentset{\mbox{\large\bfseries .}}{#1}}
\providecommand{\dx}[1][]{%
   \ifthenelse{ \equal{#1}{} }
      {\bigdot{\mathbf{x}}}
      {\bigdot{\mathbf{x}}^{[#1]}}}
\providecommand{\dxb}[1][]{%
   \ifthenelse{ \equal{#1}{} }
      {\bigdot{\mathbf{x}}^{\rm b}}
      {\bigdot{\mathbf{x}}^{{\rm b}[#1]}}}
\providecommand{\dxa}[1][]{%
   \ifthenelse{ \equal{#1}{} }
      {\bigdot{\mathbf{x}}^{\rm a}}
      {\bigdot{\mathbf{x}}^{{\rm a}[#1]}}}

\providecommand{\hofx}[1][]{%
   \ifthenelse{ \equal{#1}{} }
      {\mathbf{z}}
      {\mathbf{z}^{[#1]}}}
\providecommand{\hofxb}[1][]{%
   \ifthenelse{ \equal{#1}{} }
      {\mathbf{z}^{\rm b}}
      {\mathbf{z}^{{\rm b}[#1]}}}
\providecommand{\hofxa}[1][]{%
   \ifthenelse{ \equal{#1}{} }
      {\mathbf{z}^{\rm a}}
      {\mathbf{z}^{{\rm a}[#1]}}}
\providecommand{\dhofxb}[1][]{%
   \ifthenelse{ \equal{#1}{} }
      {\bigdot{\mathbf{z}}^{\rm b}}
      {\bigdot{\mathbf{z}}^{{\rm b}[#1]}}}
\providecommand{\dhofxa}[1][]{%
   \ifthenelse{ \equal{#1}{} }
      {\bigdot{\mathbf{z}}^{\rm a}}
      {\bigdot{\mathbf{z}}^{{\rm a}[#1]}}}
\renewcommand{\d}[1][]{%
   \ifthenelse{ \equal{#1}{} }
      {\mathbf{d}}
      {\mathbf{d}^{[#1]}}}

\providecommand{\errb}[1][]{%
   \ifthenelse{ \equal{#1}{} }
      {\varepsilon^{\rm b}}
      {\varepsilon^{{\rm b}[#1]}}}
\providecommand{\erra}[1][]{%
   \ifthenelse{ \equal{#1}{} }
      {\varepsilon^{\rm a}}
      {\varepsilon^{{\rm a}[#1]}}}
\providecommand{\erro}[1][]{%
   \ifthenelse{ \equal{#1}{} }
      {\varepsilon^{\rm obs}}
      {\varepsilon^{{\rm obs}[#1]}}}

\providecommand{\errm}[1][]{%
   \ifthenelse{ \equal{#1}{} }
      {\eta}
      {\eta^{[#1]}}}
\providecommand{\wb}[1][]{%
   \ifthenelse{ \equal{#1}{} }
      {w^{\rm b}}
      {w^{{\rm b}[#1]}}}
\providecommand{\wa}[1][]{%
   \ifthenelse{ \equal{#1}{} }
      {w^{\rm a}}
      {w^{{\rm a}[#1]}}}

\providecommand{\dd}{\mathrm{d}}    

\def\!#1{\mathcal{#1}}
\def\*#1{\boldsymbol{\mathbf{#1}}}
\DeclareMathOperator*{\argmin}{arg\,min}

\usepackage{booktabs}

\begin{document}





\title{Ensemble based Closed-Loop Optimal Control using Physics-Informed Neural Networks}





\cslauthor{Jostein Barry-Straume,\\Adwait D. Verulkar, Arash Sarshar, \\Andrey A. Popov, Adrian Sandu}

\cslyear{25}
\cslreportnumber{7}
\cslemail{jostein@vt.edu}

\csltitlepage{}

\begin{abstract}
The objective of designing a control system is to steer a dynamical system with a control signal, guiding it to exhibit the desired behavior.
The Hamilton-Jacobi-Bellman (HJB) partial differential equation offers a framework for optimal control system design. However, numerical solutions to this equation are computationally intensive, and analytical solutions are frequently unavailable. 
%
Knowledge-guided machine learning methodologies, such as physics-informed neural networks (PINNs), offer new alternative approaches that can alleviate the difficulties of solving the HJB equation numerically.
This work presents a multistage ensemble framework to learn the optimal cost-to-go, and subsequently the corresponding optimal control signal, through the HJB equation.
Prior PINN-based approaches rely on a stabilizing the HJB enforcement during training. Our framework does not use stabilizer terms and offers a means of controlling the nonlinear system, via either a singular learned control signal or an ensemble control signal policy. 
Success is demonstrated in closed-loop control, using both ensemble- and singular-control, of a steady-state time-invariant two-state continuous nonlinear system with an infinite time horizon, accounting of noisy, perturbed system states and varying initial conditions.
\end{abstract}








\section{Introduction}
\label{sec:Introduction}
The goal of optimal control is to design a control policy that guides a system from an initial state to a desired final state while minimizing the cost associated with the transition \citep{kirk2004optimal, brunton_kutz_2019}. Classical optimal control methods perform well primarily with linear systems \citep{kirk2004optimal}. 
For nonlinear systems the Hamilton-Jacobi-Bellman (HJB) partial differential equation (PDE) provides a robust framework for designing optimal control systems.
It integrates the control policy into a cost function and minimizes it over all intermediate steps. This ensures that each segment of the overall state trajectory satisfies the Bellman Optimality principle \citep{brunton2022nonlinearcontrol}. 
However, analytical solutions to the HJB equation are only available for a limited class of nonlinear differential equations \citep{khalil2002nonlinear}, and numerical solutions are computationally expensive \citep{Geering_2007}.
Classical numerical algorithms scale exponentially with the number of variables, making them susceptible to the curse of dimensionality.
Knowledge-guided machine learning (KGML) methodologies, such as physics-informed neural networks (PINNs), offer promising alternatives for addressing the challenges of solving the HJB equation for nonlinear problems.

KGML addresses both scientific problems and traditional machine learning tasks by leveraging empirical data and prior knowledge \citep{hao2022physics}. Within this framework, physics-informed neural networks (PINNs) solve semi-supervised learning tasks while respecting the governing physical laws \citep{raissi2019physics}. This is typically achieved by encoding the mathematical equations governing the physical system into the loss function.

This work introduces a multistage ensemble framework to learn the optimal cost-to-go and the corresponding optimal control signal by solving the HJB equation for a time-invariant, continuous nonlinear system with an infinite time horizon.
Previous approaches have relied on stabilization techniques to enforce the HJB equation during training \citep{dierks2009optimal,dierks2010optimal, zargarzadeh2014adaptive}. Other prior approaches removed the need for a boundary condition loss component \citep{furfaro2022physics}. Our framework eliminates the need for stabilizer terms and enables control of nonlinear systems through either a singular learned control signal or an ensemble control signal policy.
We demonstrate success in closed-loop control scenarios using both ensemble and singular control methods. Good performance is maintained in the presence of noisy, perturbed system states and varying initial conditions.
Moreover, we develop an adaptive control policy based on averaging an ensemble of control policies, excluding outlier control signals while seeking to reincorporate them into the overall mean signal when their respective control converges back to the mean.

The remainder of the paper is organized as follows. \Cref{sec:Survey} surveys approaches to optimal control using KGML methods. \Cref{sec:Methodology} provides background on optimal control problems and the HJB equation.
The novel ensemble Physics-Informed Neural Network (PINN) framework is introduced in \Cref{subsec:Time-invariant-case}.
 \Cref{sec:Experiment} presents the experimental results. Finally, \Cref{sec:Conclusions} summarizes the contributions of the paper and discusses potential directions for future work.

\section{Optimal Control with Machine Learning}
\label{sec:Survey}

Optimal control has been extensively studied in the context of machine learning, yielding a variety of innovative approaches. This section reviews several key contributions to the field.

\subsection{Reinforcement Learning in Optimal Control}
Reinforcement learning (RL) has emerged as a powerful framework for solving dynamic and stochastic control problems, particularly in environments where traditional optimization techniques struggle. RL algorithms, such as Q-learning and Proximal Policy Optimization (PPO), enable agents to learn optimal policies through trial and error \citep{sutton2018reinforcement}. Applications include robotic path planning, autonomous vehicles, and energy management \citep{lillicrap2015continuous, silver2016mastering}.

However, RL methods often suffer from high sample complexity and instability during training, particularly in high-dimensional or partially observable environments. To mitigate these challenges, recent research integrates RL with model-based approaches, leveraging domain knowledge to guide exploration and policy refinement \citep{chua2018deep}.

\subsection{Knowledge-Guided Machine Learning for Control}
Knowledge-guided machine learning (KGML) combines data-driven methods with physics-based models to address the limitations of purely empirical approaches. KGML frameworks have been particularly effective in scenarios involving complex dynamical systems, where they exploit prior knowledge to constrain solutions \citep{hao2022physics, willard2022integrating,raj2025deepoperator}.

Physics-informed neural networks (PINNs) are a notable KGML method that incorporates governing mathematical equations, such as the Hamilton-Jacobi-Bellman (HJB) equation, directly into the training process \citep{raissi2019physics,Barry-Straume2022May}. PINNs enable the solution of high-dimensional control problems while ensuring physical consistency, with applications spanning fluid dynamics, structural analysis, and optimal trajectory planning \citep{karniadakis2021physics}.

\subsection{Ensemble-based Methods and Robust Control}
Ensemble-based methods enhance the robustness and generalizability of machine learning models for control tasks. By combining predictions from multiple models, ensemble methods reduce variance and improve stability in dynamic systems \citep{dietterich2000ensemble}.

In \cite{agrachev2022control}, the authors study the \textquotedblleft problems of ensemble controllability of the control systems and the action of the flows, generated by the control systems, on the manifold of mappings.\textquotedblright{} A strength of their approach is presenting multiple examples of nonlinear control systems that demonstrate approximate controllability properties for diffeomorphisms, toruses, and two-dimensional spheres. By the same token, the authors in \cite{Staritsyn2022} generate ensemble feedback controls of a system of local continuity equations via a deterministic optimization technique using Pontryagin's Maximum Principle. The foundational idea behind the approach in \citep{Staritsyn2022} is combining the averaging principle from multi-agent systems control with the feedback control variations technique involving Pontryagin's maximum principle.

Of similar kind is the work presented in \cite{ChilledWater2021}, in which the authors use ensemble learning methods for the cooling load prediction of a chilled water system. The foundational idea of the approach in \citep{ChilledWater2021} is learning, in real time, the optimal control strategy using a cloud edge terminal form. In a parallel fashion, the authors of \cite{lee2019ensemble} present an ensemble of Bayesian Neural Networks (BNNs) for control of an autonomous driving task. The foundational idea behind the approach in \citep{lee2019ensemble} is using uncertainty quantification via an ensemble of BNNs for hypothesis testing to guide a car around a track in the event of a sensor failure. Likewise, the authors in \cite{agrachev2016ensemble} explore controlling an ensemble of nonlinear control systems using only a single parameter-independent control. They leverage Lie algebraic methods to lay the groundwork for generality of exact controllability properties for finite ensembles, and then present a three-dimensional model problem that is approximately controllable. Additionally, the authors in \cite{Beauchard_2010} study the ensemble controllability for a prototype of a infinite dimensional bilinear continuous system whose control is unknown.

\subsection{Infinite Horizon Problems}
In \cite{fotiadis2023physics}, the authors circumvent the non-unique solution to the infinite horizon optimal control problem by solving the HJB equation with a PINN for the finite horizon case, which is subsequently used as an approximator for the infinite horizon. Verification that the approximator has a sufficiently large enough finite horizon is shown. In the same vein, the authors of \cite{furfaro2022physics} improve upon the standard PINN framework by leveraging the Theory of Functional Connections. Their framework, PINN-TFC, solves several nonlinear infinite horizon optimal control problems. The main benefit to PINN-TFC is removing the need for a boundary condition component in the loss function.

Building upon this work, \cite{schiassi2022bellman} present Bellman neural networks to solve both finite- and infinite-horizon optimal control problems with integral quadratic cost with linear and nonlinear system dynamics.

\subsection{Actor-Critic and Dual Learning Approaches}
In \cite{dey2024data}, the authors present a dual learning framework for nonlinear receding horizon systems that is situation aware and offers safety guaranteed control. Their framework's main benefit is combining the idea of an actor-critical neural network together with those of optimal control perturbation theory to learn both the slow and fast variables of a nonlinear system \citep{khalil2002nonlinear}.

Another dual learning framework is presented in \cite{nishimura2024combined}, in which the authors present their Uncertain Control Co-design framework, which optimizes both the plant and controller design simultaneously. Their framework uses an alpha scaling term for both the HJB loss component, as well as the boundary condition component, for their PINN loss function. In a similar fashion, the authors of \cite{mukherjee2023actor} develop an actor-critic framework to control a one-dimensional quasi-linear battery pack cooling problem using a PINN to solve the HJB. Their framework incorporates PINN elements into a reinforcement learning framework to achieve optimal control policies that resemble Q-learning. Along the same lines, the authors in \cite{dierks2009optimal} solve the HJB equation for an affine nonlinear discrete-time system with an action neural network (NN) and critic NN framework. The former produces near optimal control signals, and the latter evaluates the former's performance, which in turn is used to update both NNs' weights.

\subsection{Challenges and Future Directions}
Despite significant advancements, several challenges remain in the integration of machine learning with optimal control. Key issues include the computational cost of training complex models, the need for large labeled datasets, and the difficulty of ensuring stability and robustness in real-world applications \citep{brunton_kutz_2019}. Future work may focus on hybrid approaches that combine the strengths of RL, KGML, and ensemble methods, along with advancements in explainability and interpretability of learned control policies.

The novel aspects of this work include an outlier excluding ensemble mean control signal policy, the non-reliance of the averaging principle from multi-agent systems, and solving of the HJB equation for the infinite time horizon case optimally without relying on sufficient approximation via finite horizon. Success of the proposed methodology is demonstrated via a 2-state continuous nonlinear affine system. Three control policies are put forth in this work: (1) ensemble control with outlier inclusion, (2) ensemble control with outlier exclusion, and (3) individual control.

\section{Optimal closed-loop control}
\label{sec:Methodology} 

To provide a rigorous foundation for our proposed method, the following subsections review the mathematical formulation of closed-loop control problems (\cref{subsec:General-Case-Problem-Setup}), an exploration of Bellman’s Principle of Optimality and an explanation of the Hamilton-Jacobi-Bellman (HJB) equation as a pivotal tool in solving continuous-time optimal control problems (\cref{subsec:Hamilton-Jacobi-Bellman-equation}). Furthermore, Pontryagin’s Maximum Principle and the Hamiltonian function are introduced to elucidate their connection to the HJB equation (\cref{subsec:Hamiltonian}).
For readers seeking an in-depth understanding of nonlinear optimal control, comprehensive treatments can be found in \citep[Ch.3, p.75]{Geering_2007} and \citep[Ch.12, p.469]{khalil2002nonlinear}.
\subsection{The Framework for Optimal Control}
\label{subsec:General-Case-Problem-Setup}

An optimal control problem seeks to minimize a cost functional over time, subject to a system of differential equations that govern the system dynamics. Formally, the system dynamics are described by:
\begin{equation}
    \label{eqn:problem-setup}
    \dot{x}(t) = f\bigl(x(t), u(t)\bigr), \quad x(t_0) = x_0, \quad t \in [t_0, t_f], \quad x(t) \in \Omega \subset \mathbb{R}^n, \quad u(t) \in \mathbb{R}^m,
\end{equation}
were \(x(t)\) is the state vector at time \(t\), \(u(t)\) is the control vector, \(\dot{x}(t)\) denotes the time derivative capturing the rate of change of the state \(x(t)\), and \(f(x, u)\) represents the system's dynamics. 
We assume that there is a compact solution domain $\Omega \subset \mathbb{R}^n$ such that the controlled dynamics \eqref{eqn:problem-setup} leads to solutions that remain within this domain, $x(t) \in \Omega$ for all $t \in [t_0, t_f]$.

Consider the following \emph{unoptimized} cost functional:
\begin{equation}
\label{eqn:unoptimized-cost-to-go}
\mathcal{J}(x,u) = \int_{t_0}^{t_f} \mathcal{L}\bigl(x(t'), u(t')\bigr) \, \dd t' + \mathcal{L}_f(x(t_f)),
\end{equation}
where \(\mathcal{L}\bigl(x(t'), u(t')\bigr)\) represents the running cost incurred over the trajectory, and \(\mathcal{L}_f(x(t_f))\) is the terminal cost at the final state. The total cost-to-go, \(\mathcal{J}\), encapsulates both immediate and terminal contributions. It is worth noting that \cref{eqn:unoptimized-cost-to-go} depends on the system state $x$, as well as the control signal $u$. This makes sense, because changing the initial conditions of the system state $x$ will impact the running total cost $\mathcal{L}(\cdot)$ in addition to the cost at final terminal state $\mathcal{L}_f(x(t_f))$. By the same token, even if the initial state condition $x(t_0)=x_0$ was held constant, one could enact different control policies resulting in varying state trajectories, and consequently different running total costs $\mathcal{L}(\cdot)$. In other words, for any prescribed control sequence, the cost-to-go $\mathcal{J}$ is a function of both $x$ and $u$ \emph{before} optimization.

The optimal control problem is to determine the control function \(u^*(t)\) for $t \in [t_0, t_f]$ that minimizes the cost functional \eqref{eqn:unoptimized-cost-to-go} subject to the model dynamics \eqref{eqn:problem-setup}:
\begin{equation}
\label{eqn:optimal-control}
u^* = \underset{u}{\argmin}\; \mathcal{J}(x,u) \quad \textnormal{subject to } \eqref{eqn:problem-setup}.
\end{equation}

To solve the optimal control problem \eqref{eqn:optimal-control}  the following \textit{value function} \(V(t, x)\) is introduced, which captures the minimum cost-to-go from a given state \(x(t) = x\) at time \(t\) under an optimal control policy $u(t')$, $t \le t' \le t_f$:
\begin{equation}
\label{eqn:value-function}
V(t, x) = \underset{u}{\min} \left[ \int_{t}^{t_f} \mathcal{L}\bigl(x(t'), u(t')\bigr) \, \dd t' + \mathcal{L}_f(x(t_f)) \right].
\end{equation}
The value function \eqref{eqn:value-function} plays a central role in optimal control, as it formalizes the goal of minimizing the total cost-to-go. By solving for \(u(t')\) that minimizes \(V(t, x)\), we derive the \textit{optimal control law}. This control law ensures that the system's trajectory \(x(t')\), $t \le t' \le t_f$, minimizes the total cost functional \(\mathcal{J}\).

\begin{remark}[Key distinctions between Cost-to-Go ($\mathcal{J}$) and Value ($V$) functions]
$\mathcal{J}$ is linked to a specific control, whereas $V$ is the optimized cost. $\mathcal{J}$ measures what the total cost actually is if one uses a particular control $u(t)$ from state $x$. $V$ measures the smallest possible cost under any control from state $x$. Furthermore, a given control $u(t)$ results in a scalar cost $\mathcal{J}$, whereas the value function $V$ is usually an unknown function that must be computed by minimizing over all feasible controls. In summary, $V$ is derived by taking the minimum over all possible $\mathcal{J}$ values \citep{bryson2018applied, sontag2013mathematical, evans2022partial}.
\end{remark}

Dynamic programming (DP) provides a framework for solving control problems \eqref{eqn:optimal-control} by breaking them into smaller subproblems that are solved recursively \citep{bellman1957dynamic, bertsekas1996dynamic}. This recursive nature is captured by the following.

\begin{definition}[Bellman’s Principle of Optimality (PO)]
    \label{def:Principle-of-Optimality}
    An optimal policy has the property that, regardless of the initial state and initial decision, the remaining decisions must constitute an optimal policy with respect to the state resulting from the first decision \citep{bellman1957dynamic}.
\end{definition}

The Principle of Optimality allows the value function \eqref{eqn:value-function}  to be computed recursively, ensuring that optimal decisions at each stage minimize both the immediate and future costs. This is formalized in discrete time as follows.
Let the time horizon be discretized as 
\[
t_{i} = t_0 + i \,\frac{t_f - t_0}{M}, \quad i = 0, \dots, M. 
\]
At the final time step \(t_f\), the value function $V$ of the cost-to-go $\mathcal{J}$ is equal to the terminal cost:
\begin{equation}
    \label{eqn:PO-base-case}
    V(t_f, x_{t_f}) = \mathcal{L}_f(x_{t_f}).
\end{equation}
At any earlier time step \(t_{i}\), the value function is computed recursively from value function at later time steps:
\begin{equation}
    \label{eqn:PO-recursive}
    V(t_{i}, x_{t_i}) = \underset{u_{t_{i}}}{\min} \Biggr[ \mathcal{L}(x_{t_{i}}, u_{t_{i}}) + V(t_{i+1}, x_{t_{i+1}}) \Biggr].
\end{equation}

Given \cref{def:Principle-of-Optimality} and \cref{eqn:PO-recursive}, the optimized cost-to-go $\mathcal{J}^*$ is thus:

\begin{equation}
    \label{eqn:optimized-cost-to-go}
    \mathcal{J}^{*}(t,x)\;=\;\inf_{u(\cdot)} \mathcal{J}\bigl(t,x;u(\cdot)\bigr).
\end{equation}

\begin{remark}[Dependent variables of the Cost-to-Go $\mathcal{J}$ before and after optimization]
    It is important to note the change in dependent variables between \cref{eqn:unoptimized-cost-to-go} and \cref{eqn:PO-recursive}, and \cref{eqn:optimized-cost-to-go}. Again, before optimization, \cref{eqn:unoptimized-cost-to-go} depends on the system state $x$ and the control signal $u$. However, \emph{after optimization occurs}, \cref{eqn:PO-recursive} is a scalar field on state space only since the control dependency has been absorbed by minimization. In other words, the control signal $u$ no longer appears as an explicit term. The value function $V$, being the optimized cost of all possible cost-to-go $\mathcal{J}$'s in \cref{eqn:optimized-cost-to-go}, is a function of the system state $x$ only (and time $t$ if appropriate). This makes sense, as the optimal control signal $u^*$ has been found via recursively minimizing the running total cost $\mathcal{L}(\cdot)$. Said optimal control policy that corresponds to \cref{eqn:PO-recursive} can be recovered by using the HJB equation or Pontryagin's Principle, but it is no longer a dependent argument for the value function $V$, and therefore the cost-to-go $\mathcal{J}$'s that comprise it.

    \begin{table}[]
    \centering
    \caption{Dependent variables of Cost-to-Go $\mathcal{J}$ before and after optimization}
    \begin{tabular}{@{}llp{7.5cm}@{}}
    \toprule
    Stage & Symbol & Free arguments \\
    \midrule
    Pre‑optimization & $\mathcal{J}(t,x;u(\cdot))$ & Current state $x$ and candidate control trajectory $u(\cdot)$ \\
    Post‑optimization & $\mathcal{J}^{*}(t,x)$ & Current state (and $t$ if finite‑horizon); optimal policy is implicit \\
    \bottomrule
    \label{tab:cost-to-go-dependent-variables}
    \end{tabular}
    \end{table}
\end{remark}

\subsection{Hamilton-Jacobi-Bellman (HJB) Equation}
\label{subsec:Hamilton-Jacobi-Bellman-equation}

The Hamilton-Jacobi-Bellman (HJB) equation \citep{bellman1957dynamic} formalizes the recursive structure of dynamic programming in a continuous-time setting, enabling the derivation of optimal control policies. The principle of optimality shows us that for the entire trajectory to be optimal one can enforce optimality over infinitesimal time increments. Over a small time interval \(\Delta t\), the value function \cref{eqn:PO-recursive}  can be expressed as:

\begin{equation}
    \label{eqn:V-taylor-expanded}
    V(t, x) = \underset{u(t)}{\min} \Biggr[ \mathcal{L}\bigl(x(t), u(t)\bigr) \Delta t + V(t + \Delta t, x(t + \Delta t)) \Biggr].
\end{equation}
Expanding \(V(t + \Delta t, x(t + \Delta t))\) in a Taylor series around \(t\) yields:
\begin{equation}
    \label{eqn:V-taylor-series}
    V(t + \Delta t, x(t + \Delta t)) \approx V(t, x) + \dot{V}(t, x) \Delta t + \nabla_x V \cdot f\bigl(x(t), u(t)\bigr) \Delta t,
\end{equation}
where \(\dot{V}(t, x)\) is the time derivative of the value function, and \(\nabla_x V \cdot f(x, u)\) is the inner product of the gradient of \(V\) with the system dynamics \(f(x, u)\). Substituting \cref{eqn:V-taylor-series} into \cref{eqn:V-taylor-expanded} and subtracting \(V(t, x)\) from both sides gives:
\begin{equation}
    \label{eqn:V-after-expansion}
    0 \approx \underset{u(t)}{\min} \Biggr[ \mathcal{L}\bigl(x(t), u(t)\bigr) + \dot{V}(t, x) + \nabla_x V \cdot f\bigl(x(t), u(t)\bigr) \Biggr] \Delta t.
\end{equation}
Taking the limit as \(\Delta t \to 0\), the HJB equation emerges:
\begin{equation}
    \label{eqn:hjb-equation}
    \dot{V}(t, x) + \underset{u(t)}{\min} \Biggr[ \mathcal{L}\bigl(x(t), u(t)\bigr) + \nabla_x V \cdot f\bigl(x(t), u(t)\bigr) \Biggr] = 0.
\end{equation}
The HJB \cref{eqn:hjb-equation} can then be rewritten as:
\begin{equation}
    \label{eqn:hjb-with-hamiltonian}
    \dot{V}(t, x) + \underset{u}{\min} \, \mathcal{H}(x, u, \nabla_x V) = 0,
\end{equation}
with the minimization term in \cref{eqn:hjb-equation} referred to as the Hamiltonian function:
\begin{equation}
    \label{eqn:hamiltonian-definition}
    \mathcal{H}(x, u, \nabla_x V) = \mathcal{L}(x, u) + \nabla_x V \cdot f(x, u).
\end{equation}
The optimal control \(u^*(t)\) minimizes the Hamiltonian such  that
\begin{equation}
    \label{eqn:optimal-control-hjb}
    u^*(t) = \underset{u}{\argmin} \, \mathcal{H}(x(t), u(t), \nabla_x V).
\end{equation}
\begin{remark}[Steady-state HJB]
\label{sec:remark1}
For time-invariant systems, the value function \(V(x)\) depends only on the state \(x\), and the time derivative \(\dot{V}(t, x)\) vanishes. The HJB equation \eqref{eqn:hjb-equation} simplifies to:

\begin{equation}
    \label{eqn:steady-state-hjb}
    \underset{u}{\min} \Biggr[ \mathcal{L}(x, u) + \nabla_x V \cdot f(x, u) \Biggr] = 0.
\end{equation}

This steady-state form is particularly relevant for infinite-horizon problems, where the objective is to minimize long-term costs and ensure system stability. Solving \cref{eqn:steady-state-hjb} means selecting the best control signal $u$ for each system state $x$. Once the steady-state HJB equation is solved, the best control signal $u$ is then already selected, and the control trajectory becomes an implicit variable.
\end{remark}

\subsection{Pontryagin's Principle and the Hamiltonian}
\label{subsec:Hamiltonian}

Pontryagin’s Principle provides necessary conditions for optimality in continuous-time control problems. It reformulates the control  problem as a boundary value problem involving the system state \(x(t)\), the control input \(u(t)\), and an auxiliary variable known as the costate \(\lambda(t)\). This approach complements the recursive framework of the HJB equation by providing an alternative perspective on optimality.

The cost functional \eqref{eqn:unoptimized-cost-to-go} is augmented by introducing a Lagrange multiplier \(\lambda(t)\) to enforce the system dynamics \(\dot{x} = f(x, u, t)\) as a constraint, and building the following Lagrangian functional:
\begin{equation}
    \label{eqn:augmented-cost}
    \mathcal{C} = \mathcal{J} + \int_{t_0}^{t_f}  \lambda^\top \big( \dot{x} - f(x, u, t) \big)  \, \dd t.
\end{equation}
To find the necessary conditions for optimality, derivatives of \(\mathcal{C}\) with respect to \(x(t)\), \(\lambda(t)\), and \(u(t)\) are equated to zero. The resulting conditions can be summarized using the Hamiltonian function, here defined as:
\begin{equation}
    \label{eqn:pontryagin-hamiltonian}
    \mathcal{H}(x, \lambda, u, t) = \mathcal{L}(x, u, t) + \lambda^\top f(x, u, t).
\end{equation}
The necessary conditions for optimality are given by \citep{pontryagin2018mathematical}:
\begin{subequations}
    \label{eqn:pontryagin-conditions}
    \begin{align}
        \dot{x}(t) \quad &= \quad \frac{\partial \mathcal{H}}{\partial \lambda}, \quad  \text{(State equation),} \label{eqn:state-dynamics} \\
        \dot{\lambda}(t) \quad &= -\frac{\partial \mathcal{H}}{\partial x}, \quad \text{(Costate equation),} \label{eqn:costate-dynamics} \\
        \frac{\partial \mathcal{H}}{\partial u} \quad &= \quad 0, \quad  \text{(Stationarity condition).} \label{eqn:stationarity-condition}
    \end{align}
\end{subequations}
The boundary conditions are \(x(t_0) = x_0\) and \(\lambda(t_f) = \nabla \mathcal{L}_f(x(t_f))\). Together, these equations define a two-point boundary value problem that characterizes the optimal trajectories of both the state and the costate.
\Cref{eqn:stationarity-condition} indicates that the optimal control $u^{*}(t)$ is the stationary point of the Hamiltonian. 
If $\lambda(t) = \nabla_x V(x,t)$ then both HJB and Pontryagin's principle lead to the optimal control \(u^*(t)\)  minimizing the Hamiltonian:
\begin{equation}
   \label{eqn:pontryagin-control}
u^*(t) = \underset{u}{\argmin} \, \mathcal{H}(x, u, \lambda, t) = \underset{u}{\argmin} \, \mathcal{H}(x, u, \nabla_x V, t).
\end{equation}

Note that the control signal $u^*$ is an output of the optimization rather than an argument of \cref{eqn:pontryagin-control}. In the next \Cref{subsec:Time-invariant-case}, the proposed ensemble physics-informed architecture will leverage Bellman's ideas with Pontryagin's notations to solve the steady-state time-invariant control problem with an infinite time horizon.
    
%

\section{Solving Time-Invariant Control Problems with an Ensemble of Physics-Informed Neural Networks (PINNs)}
\label{subsec:Time-invariant-case}

\subsection{Time-Invariant Affine Control Problems with Quadratic costs}
\label{subsec:affine-case}

We now focus on autonomous systems \eqref{eqn:problem-setup} with control-affine system dynamics:
\begin{equation}
\label{eqn:affine-system}
\dot{x}(t) = f(x, u) = f_1(x) + f_2(x) u,
\end{equation}
where \(f_1(x)\) captures the state-dependent dynamics, and \(f_2(x)\) describes the effect of the control on the dynamics.  Both \(f_1\) and \(f_2\)  are nonlinear functions.

For the control goals we consider a quadratic running cost \eqref{eqn:unoptimized-cost-to-go}:
\begin{equation}
\label{eqn:quadratic}
\mathcal{L}(x, u) = x^\top\, Q\, x + u^\top\, R\, u,
\end{equation}
where \(Q\) and \(R\) are positive definite matrices that penalize state and control signals respectively. A quadratic cost is a widely used choice in control theory since the convex loss landscape allows for closed-form analytical solutions which can serve as benchmarks for validating solutions in nonlinear settings.  In addition, we consider a zero terminal cost $\mathcal{L}_f \equiv 0$ \eqref{eqn:unoptimized-cost-to-go}.

Our goal is to solve the optimal control problem \eqref{eqn:optimal-control} {\it over the infinite time horizon} $t_0 = \tau = 0$, $t_f \to \infty$ for the particular choices \eqref{eqn:affine-system} and \eqref{eqn:quadratic}. Formally, the optimal control problem treated here is defined as:
\begin{subequations}
    \label{eqn:time-invariant-control-problem}
    \begin{eqnarray}
    \label{eqn:time-invariant-control-problem-min}
 &&   \underset{u}{\min} ~~ \int^{\infty}_{\tau} \left(x^\top(t')\, Q\, x(t') + u^\top(t')\, R\, u(t')\right) \, \dd t',\\
    \label{eqn:time-invariant-control-problem-ode}
 &&   \text{subject to} \quad \dot{x}(t) = f_1\bigl(x(t)\bigr) + f_2\bigl(x(t)\bigr)\, u(t), \quad x(\tau) = x_\tau, 
    \end{eqnarray}
\end{subequations}
Substituting \eqref{eqn:affine-system} and \eqref{eqn:quadratic} into \cref{eqn:steady-state-hjb} leads to the following steady-state HJB equation:
\begin{equation}
    \label{eqn:hjb-time-invariant}
    \underset{u}{\min} \Biggr [ x^\top\, Q\, x + u^\top\, R\, u + \nabla_x \mathcal{J} \cdot \left( f_1(x) + f_2(x) u \right) \Biggr ] = 0.
\end{equation}
Substituting \eqref{eqn:affine-system} and \eqref{eqn:quadratic} into \eqref{eqn:hamiltonian-definition} leads to the Hamiltonian:
\begin{equation}
    \label{eqn:hamiltonian-expanded}
    \mathcal{H}(x, u, \lambda) = \underbrace{x^\top\, Q\, x}_{\text{state cost}} + \underbrace{u^\top\, R\, u}_{\text{control cost}} + \underbrace{\lambda^\top \big( f_1(x) + f_2(x) u}_{\text{costate term}} \big).
\end{equation}
The optimal control \(u(t)\) \eqref{eqn:optimal-control-hjb}  minimizes the Hamiltonian \eqref{eqn:hamiltonian-expanded} and is defined as
\begin{equation}
    u= -\frac{1}{2} \,R^{-1}\, f^T_2(x)\, \Bigl( \frac{\partial \mathcal{J}}{\partial x} \Bigr)^T,
    \label{eqn:control-signal}
\end{equation}

where $\frac{\partial \mathcal{J}}{\partial x}$, which is equivalently known as the costate $\lambda$, is the derivation of the cost-to-go $\mathcal{J}$ with respect to the system state $x$. We assume that there is a compact solution domain $\Omega \subset \mathbb{R}^n$ such that the optimally controlled dynamics \eqref{eqn:time-invariant-control-problem} leads to solutions that remain within this domain, $x(t) \in \Omega$ for all $t \ge \tau$.

\subsection{Discrete-time closed loop control strategy}
\label{subsec:discrete-time}

We seek to perform closed-loop control the system \eqref{eqn:problem-setup}--\eqref{eqn:affine-system} during its evolution for $t \ge t_0$.  For this we consider discrete times
\[
t_0,\quad t_1 = t_0 + \Delta t, \quad \dots, \quad t_m = t_0 + m\,\Delta t, \dots
\]
Denote by $x_{m}$ the system state at time $t_m$. We compute the instantaneous control signal value $u_{m}$ at $t_m$ as follows:
\begin{itemize}
\item Solve the optimal control problem \eqref{eqn:time-invariant-control-problem} for the initial time $\tau = t_m$ and starting from the initial state $x_\tau = x_{m}$. The result is optimal control function $u^*(t)$ for $t \ge t_m$.
\item Set the current control signal to be the optimal control value at the beginning of the interval: $u_{m} = u^*(t_m)$.
\end{itemize}
This control strategy is closed-loop since the control signal $u_m$ depends on the system state $x_m$.

Next, the system state is advanced from $t_m$ to $t_{m+1} =  t_{m}+\Delta t$ using a discretized form of the dynamical equations \eqref{eqn:problem-setup}--\eqref{eqn:affine-system}, and using a constant control signal value over the $\Delta t$ interval:
   \begin{equation}
   \label{eqn:state_update}
   x_{m+1} = x_m + \Delta t \cdot f\left(x_m, u_m\right).
   \end{equation}
Now we arrived at time $t_{m+1}$ with the system state $x_{m+1}$. The process is repeated to compute the instantaneous control signal value $u_{m+1}$, and so on.

\subsection{Ensemble of Physics-Informed Neural Networks (PINNs) for Solving Control Problems}

\subsubsection{Neural Network to approximate Cost-to-Go}

A neural network can be employed to approximate the \emph{optimized} cost-to-go $\mathcal{J}^*$ as detailed in \cref{eqn:optimized-cost-to-go}, where \(\mathcal{J}^*:\mathbb{R}^n \to \mathbb{R}\), and where \(x \in \mathbb{R}^n\) denotes the system state. Let \(\widehat{\mathcal{J}}(x; \theta)\) be the neural-network approximation, with \(\theta\) representing the network’s learnable parameters (weights and biases). A standard feedforward architecture can be defined as follows.

\textbf{Input layer.} The input layer takes the state vector $x$ and treats it as

$$z^{(0)}(x) = x.$$

Here, \(z^{(0)}\in\mathbb{R}^n\) is the initial input to the network, and \(n\) is the dimensionality of the state space.

\textbf{Hidden layers.} For each hidden layer \(\ell \in \{1,\dots,L-1\}\), the transformation is  

$$
z^{(\ell)}(x;\theta) = \sigma_\ell \Bigl(W^{(\ell)},z^{(\ell-1)}(x; \theta) + b^{(\ell)}\Bigr),
$$

where \(W^{(\ell)}\in\mathbb{R}^{d_{\ell}\times d_{\ell-1}}\) and \(b^{(\ell)}\in \mathbb{R}^{d_{\ell}}\) are the weight matrix and bias vector for the \(\ell\)-th layer. \(\sigma_\ell\) denotes a nonlinear activation function (e.g., ReLU, tanh), applied elementwise. \(d_\ell\) is the dimension of layer \(\ell\)’s output.

\textbf{Output layer.} The network produces a single scalar output, since the cost-to-go function is real-valued:
\[
\widehat{\mathcal{J}}(x; \theta) 
\;=\; z^{(L)}(x; \theta)
\;=\; W^{(L)} \,z^{(L-1)}(x; \theta) \;+\; b^{(L)},
\]
where \(W^{(L)}\in \mathbb{R}^{1\times d_{L-1}}\) and \(b^{(L)}\in \mathbb{R}\). This final linear mapping translates the last hidden-layer output to a single real value.

\subsubsection{MSE Loss for sub-optimal Cost-to-Go Approximation}

The network parameters $\theta$ are adjusted (e.g., via gradient-based optimization) to minimize a loss function that enforces fidelity to the true cost-to-go. A common choice is the mean squared error (MSE), which can be written as
\begin{equation}
\mathcal{L}(\theta) 
\;=\; \frac{1}{O}\sum_{k=1}^O \Bigl(\widehat{\mathcal{J}}\bigl(x_k; \theta\bigr) - \mathcal{J}^*_{\text{true}}(\mathbf{x}_k)\Bigr)^2,
\label{eqn:MSE-loss}
\end{equation}
where $x_k$ denotes the \(k\)-th training sample, $\widehat{\mathcal{J}}\bigl(x_k; \theta\bigr)$ is the network's estimated cost-to-go, \(\mathcal{J}^*_{\text{true}}(x_k)\) is the known optimized cost-to-go (solved off-line by traditional control approaches), and $O$ is the total number of sampling points. Minimizing \(\mathcal{L}(\theta)\) encourages the neural network to accurately approximate \(\mathcal{J}\) over the sampled domain.

Once trained, the neural network serves as a surrogate model, capable of quickly evaluating the cost-to-go \(\mathcal{J}(x)\) for a wide range of states $x$. This approach can be used in settings where either an analytical expression for $\mathcal{J}$ is not available, or where one seeks to embed approximate dynamic programming or reinforcement learning concepts into a control pipeline.

\subsubsection{Boundary Data}

In a typical supervised-learning or regression setting involving a spatial domain \(\Omega \subset \mathbb{R}^n\), boundary data points (sometimes referred to as exterior data points or boundary conditions) are those samples that lie on the boundary \(\partial \Omega\). Formally, the domain and boundary can be defined by letting \(\Omega\) be a closed, bounded set in \(\mathbb{R}^n\) with boundary \(\partial \Omega\). The sample locations along the boundary are a selected finite set of points 
$\Bigl\{x_{i_b} \in \partial \Omega$, $ i_b=1,\dots,M_b\Bigr\}$.  
For each $x_{i_b}$, let $\mathcal{J}_{i_b}$ be the observed or prescribed output associated with that boundary point. The boundary training data (or exterior data points) is then  
\begin{equation}
     \mathcal{D}_b \;=\; \Bigl\{\bigl(x_{i_b}, \mathcal{J}_{i_b}\bigr)\,\big|\;x_{i_b} \in \partial \Omega, \, i_b=1,\dots,M_b\Bigr\}.
     \label{eqn:boundary-data}
\end{equation}

\subsubsection{Deriving the optimal costate and calculating the optimal control signal}

In optimal control, the costate $\lambda$ emerges as the gradient of the cost-to-go $\mathcal{J}$ with respect to the state $x$. That is,

\begin{equation}
    \lambda(x) = \nabla_x \mathcal{J}(x).
    \label{eqn:lambda}
\end{equation}

When $\mathcal{J}$ is approximated by a neural network $\widehat{\mathcal{J}}(x;\theta)$, modern automatic differentiation libraries (i.e. PyTorch) allow direct computation of $\nabla_x \widehat{\mathcal{J}}(x;\theta)$. A pseudo-code example of how to accomplish this in PyTorch is shown below.

\begin{lstlisting}[language=Python]
import torch
x = x.requires_grad_(True)
model = CostToGoNN()
J_pred = model(x)
costate = torch.autograd.grad(outputs=J_pred, inputs=x)[0]
\end{lstlisting}

With the given state $x$, the corresponding predicted cost-to-go $\widehat{\mathcal{J}}$, the derived costate $\widehat{\lambda}=\nabla_x \widehat{\mathcal{J}}$, all the necessary information is available to compute the optimal control signal $u$ via \cref{eqn:control-signal}. Furthermore, this information can be used to calculate the HJB equation left hand side \eqref{eqn:hjb-time-invariant} for the given state space. As seen in the next section, this is useful for incorporating into a learning framework's objective function, such as \cref{eqn:HJB-PINN-loss}.

\subsubsection{HJB PINN Loss for Optimal Cost-to-Go Approximation}

To adjust the network parameters $\theta$ to approximate the \emph{optimal} cost-to-go $\mathcal{J}^*$, we propose to use a HJB PINN loss function. 
In this framework, one designs a composite loss function comprised of two main components—one enforcing boundary data consistency and another enforcing HJB equation residual minimization \eqref{eqn:hjb-time-invariant}. The goal of this stage of training to further condition the network weights to better approximate the optimal cost-to-go. With this in mind, continuing to include \cref{eqn:MSE-loss} as a loss component would be inappropriate as it conditions the network weights towards sub-optimality. However, the learning task needs to be constrained while the network solves the HJB equation. Therefore, boundary data from \cref{eqn:boundary-data} is still included as a loss component in \cref{eqn:HJB-PINN-loss}. The HJB PINN loss function is defined as:

\begin{equation}
\begin{split}
\mathcal{L}(\theta) 
&= \frac{1}{O}\sum_{k=1}^O \Bigl(\widehat{\mathcal{J}}\bigl(x_{k_b}; \theta\bigr) - \mathcal{J}^*_{\text{true}}(x_{k_b})\Bigr)^2\\
&+ \alpha\, \frac{1}{O}\sum_{k=1}^O \Bigl( x_k^\top Q x_k + u_k^\top R u_k + \nabla_{x_k} \widehat{\mathcal{J}}\bigl(x_k; \theta\bigr) \cdot \big(f_1(x_k, \theta) + f_2(x_k, \theta) \widehat{u}_k\big)\Bigr)^2,
\label{eqn:HJB-PINN-loss}
\end{split}
\end{equation}
where $x_{k_b}$ are boundary sample points, $x_{k}$ are sample points containing \emph{both} interior (or collocation) and exterior (boundary) points for enforcing the HJB equation, $\widehat{\mathcal{J}}\bigl(x_{k_b}; \theta\bigr)$ is the network's estimated cost-to-go for the given sample of boundary points, \(\mathcal{J}^*_{\text{true}}(x_{k_b})\) is the known cost-to-go for the given sample of boundary points, $\nabla_{x_k} \widehat{\mathcal{J}}\bigl(x_k; \theta\bigr)$ is the estimated cost-to-go for the given sample of data points $x_k$ differentiated with respect to the said state spaces $x_k$ (i.e., the costate $\widehat{\lambda}$), $\widehat{u}_k$ is the optimal control signal which is calculated using the derived costate $\widehat{\lambda}$, $\theta$ denotes the learnable parameters in the neural network, $O$ is the total number available training data, with $k$ representing the current batch of training data, and $\alpha$ is a weighting coefficient balancing the two residual terms. The model trains on an equal amount of boundary sample points and regular sample points to balance the learning task of enforcing the HJB equation while ensuring the problem remains constrained.

The first term, the boundary data residual, enforces consistency with known or prescribed boundary conditions. Minimizing this term ensures that the neural network’s approximation \(\widehat{\mathcal{J}}(x_{k_b};\theta)\) agrees with the true cost-to-go \(\mathcal{J}^*_{\rm true}(x_{k_b})\) at the domain’s boundary.

The second term, the HJB equation component, penalizes the residual of the HJB equation. In many optimal control problems, the HJB equation expresses the condition that the instantaneous cost plus the gradient of \(\mathcal{J}\) times the system dynamics must vanish for the optimal solution. Here, \(f_1\) and \(f_2\) represent the drift and control “pass-through” terms, respectively. Minimizing this residual encourages the neural network to learn an approximation \(\widehat{\mathcal{J}}\) consistent with the HJB PDE, thus ensuring its optimality.

By jointly optimizing with respect to both boundary and interior components, one obtains a PINN-based optimal cost-to-go approximation \(\widehat{\mathcal{J}}(x;\theta)\) that meets the desired boundary constraints while satisfying the underlying optimality equation in the domain’s interior. The parameter \(\alpha\) balances the emphasis between boundary constraints and PDE residual; a larger \(\alpha\) enforces the HJB condition more aggressively, while a smaller \(\alpha\) places more emphasis on matching boundary data.

\subsubsection{Ensemble of Neural Networks}

To solve the time-invariant optimal control problem, we propose using an ensemble of Physics-Informed Neural Networks (PINNs). Each network $j \in \{1,2,\ldots,N \}$  in the ensemble of $N$ PINNs outputs approximations of the cost-to-go, denoted as \(\mathcal{\widehat{J}}^{\langle j \rangle}(x) \). The approximated adjoint variable \(\widehat{\lambda}^{\langle j \rangle} = \nabla_x \widehat{\mathcal{J}}^{\langle j \rangle}(x)\) is derived via automatic reverse differentiation on $\widehat{\mathcal{J}}^{\langle j \rangle}(x)$ with respect to the system state $x$. 

 The goal is to minimize the Hamiltonian \cref{eqn:hamiltonian-expanded} with an ensemble of physics-informed neural networks. To achieve this, we design a neural network that outputs the optimal cost-to-go ($\mathcal{J}^*$) from input data ($x$).

\begin{equation}
\label{eqn:NN-def}
[\widehat{\mathcal{J}}^{\langle j \rangle}] = \mathcal{N}^{\langle j \rangle}(x_\tau;\theta^{\langle j \rangle}), \quad j = 1,2,\ldots,N.
\end{equation}

The  optimal control \(u^{*}\) can be calculated individually per network or using statistical measures of the ensemble. We will discuss these strategies next. {\it The ensemble approach reduces the risk of relying on a single, potentially overfit network by combining the outputs of multiple models, ensuring a more reliable and robust control policy.}

The loss function for each network $j$ in the ensemble $N$ is denoted $\mathscr{L}^{\langle j \rangle}(x,\theta^{\langle j \rangle})$ where \(\theta^{\langle j \rangle}\) are the network $j$ parameters. We will discuss this loss function later.

The total ensemble loss is the ensemble average of individual network losses:
\begin{equation}
    \label{eqn:ensemble-loss}
    \mathscr{L}_\text{ensemble}(x) = \frac{1}{N} \sum_{j=1}^N \mathscr{L}^{\langle j \rangle}(x,\theta^{\langle j \rangle}).
\end{equation}
This formulation ensures that the ensemble PINNs collectively satisfy the boundary conditions and the Hamilton-Jacobi-Bellman equation, resulting in a robust approximation of the cost-to-go function.

\subsection{Strategies for Utilizing Ensemble Control}

We evaluate three strategies for applying ensemble control to the system, discussed next. 

\subsubsection{Individual Control for Each Ensemble Member}
  
   In this approach, each ensemble member generates its control signal independently. The control signal \(u^{\langle j \rangle}(x)\) predicted by the $j$-th network is applied directly to its corresponding state. The next state of each ensemble member is computed independently as:
   \begin{equation}
   \label{eqn:individual_control}
   x_{i+1}^{\langle j \rangle} = x_i^{\langle j \rangle} + \Delta t \cdot f\left(x_i^{\langle j \rangle}, u^{\langle j \rangle}(x_i^{\langle j \rangle})\right), \quad i = 1, \ldots, M, \quad j = 1, \ldots, N,
   \end{equation}
   where \(x_i\) is the state at time step \(i\), \(\Delta t\) is the time step size, and \(f(x, u)\) represents the system dynamics.
 This method ensures that each ensemble member evolves based on its own control policy.

\subsubsection{Mean Control for All Ensemble Members}
  
   In this approach, the mean control signal is calculated across all networks in the ensemble and used as the action for the next step:
   \begin{equation}
   \label{eqn:mean_control}
   \bar{u}(x) = \frac{1}{N} \sum_{j=1}^N u^{\langle j \rangle}(x),
   \end{equation}
This single control signal \(\bar{u}(x)\) is applied uniformly to all ensemble states. The next state for all ensemble members is then updated as:
   \begin{equation}
   \label{eqn:mean_control_update}
   x^{\langle j \rangle}_{i+1} = x^{\langle j \rangle}_i + \Delta t \cdot f\left(x^{\langle j \rangle}_i, \bar{u}(x_i)\right), \quad i = 1, \ldots, M, \quad j = 1, \ldots, N.
   \end{equation}

\subsubsection{Outlier Detection with Mean Control for In-Distribution Members}
  
   This strategy uses a multivariate Chauvenet criterion to detect outliers. For a dataset of ensemble states \( x^{\langle j \rangle} = \{x^{\langle 1 \rangle}, x^{\langle 2 \rangle}, \ldots, x^{\langle N \rangle}\} \), the mean vector \( \mu \) and covariance matrix \( \Sigma \) are computed:
   \begin{equation}
   \label{eqn:mean_covariance}
   \mu = \frac{1}{N} \sum_{j=1}^N x^{\langle j \rangle}, \quad
   \Sigma = \frac{1}{N-1} \sum_{j=1}^N (x^{\langle j \rangle} - \mu)(x^{\langle j \rangle} - \mu)^\top.
   \end{equation}
   The probability of each state vector \( x^{\langle j \rangle} \) is evaluated under a multivariate Gaussian distribution:
   \begin{equation}
   \label{eqn:probability_distribution}
   P(x^{\langle j \rangle}) = \frac{1}{(2\pi)^{\langle j \rangle/2} |\Sigma|^{1/2}} 
   \exp\left( -\frac{1}{2} (x^{\langle j \rangle} - \mu)^\top \Sigma^{-1} (x^{\langle j \rangle} - \mu) \right),
   \end{equation}
   and states are flagged as outliers if:
   \begin{equation}
   \label{eqn:outlier_criterion}
   P(x^{\langle j \rangle}) < C, \quad C = \frac{1}{2N}.
   \end{equation}
   For in-distribution states (\(P(x^{\langle j \rangle}) \geq C\)), the mean control signal is computed as in \cref{eqn:mean_control}. The next state for in-distribution states is then updated using \cref{eqn:mean_control_update}, while outliers are excluded from the mean calculation and are controlled individually using \cref{eqn:individual_control}.

\section{Computational Experiment}
\label{sec:Experiment}

To demonstrate the methodology proposed in this work, we consider the following time-invariant tqo-state continuous nonlinear system with an infinite time horizon from \citep{dierks2010optimal}:
\begin{equation}
\label{eqn:dierks-system}
\begin{split}
x &= \begin{bmatrix}
    x_1 & x_2
\end{bmatrix}^T,\\
b &= \cos{(2x_1)} + 2,\\
f_1(x) &= \begin{bmatrix}
    -x_1 + x_2\\
    -\frac{1}{2}\, \big(x_1 + x_2 (1 - b^2)\big)
\end{bmatrix},\\
f_2(x) &= \begin{bmatrix}
    0 & b
\end{bmatrix}^T,\\
Q(x) &= x^Tx,\\
R &= 1,
\end{split}
\end{equation}
where $x$ is the system state, $b$ is a simplified notation for a term that also defines system dynamics, $f_1(x)$ captures the state-dependent dynamics, $f_2(x)$ describes the effect of the control on the dynamics, $Q(x)$ is a positive definite matrix that penalizes system states, and $R$ is a positive definite matrix that penalizes control signals.

\subsection{Ensemble Training}

\begin{figure}
\begin{center}
  (a) \includegraphics[width=0.44\textwidth, trim=4 4 4 4,clip, keepaspectratio]{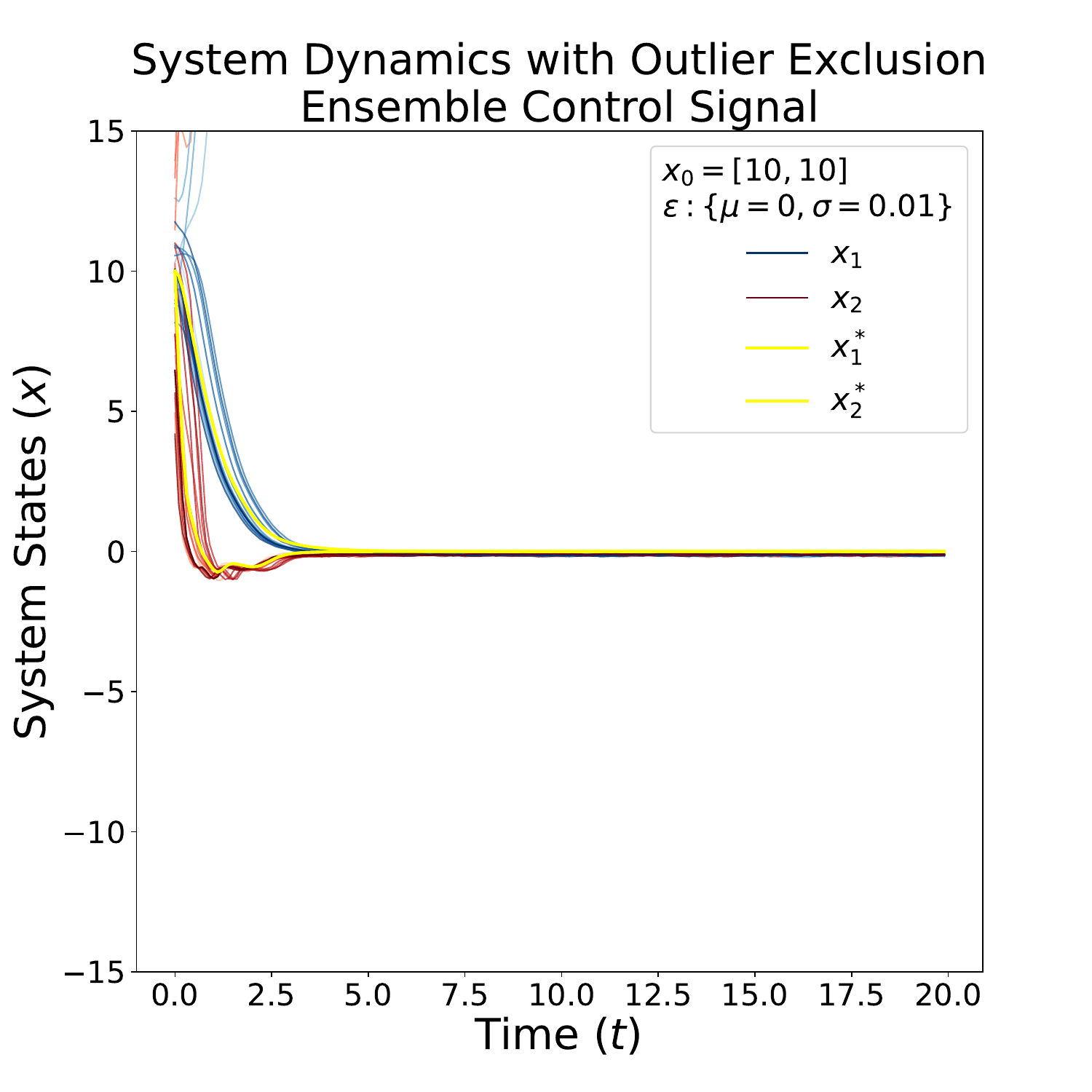}
  (b) \includegraphics[width=0.44\textwidth, trim=4 4 4 4,clip, keepaspectratio]{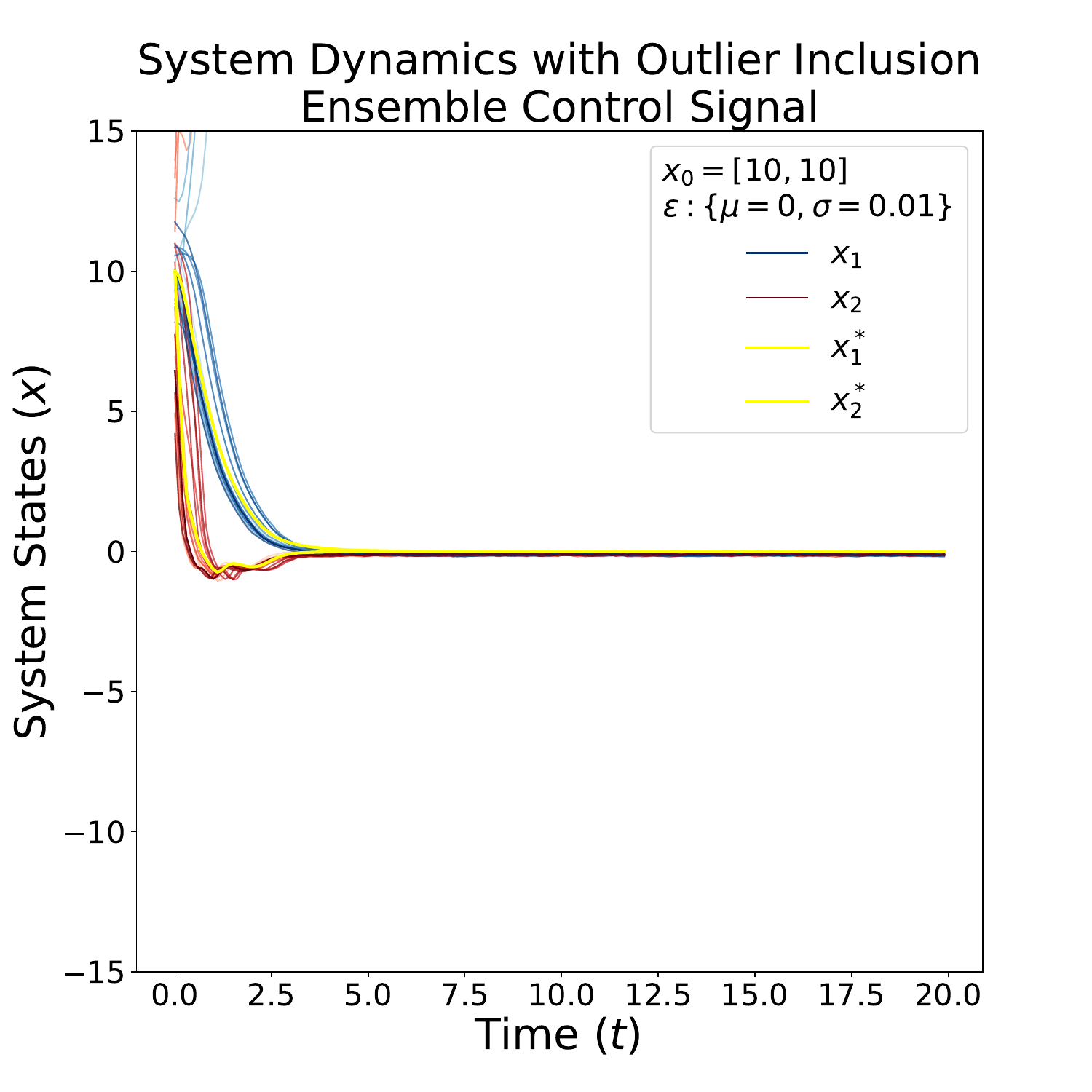}\\
  (c)  \includegraphics[width=0.44\textwidth, trim=4 4 4 4,clip, keepaspectratio]{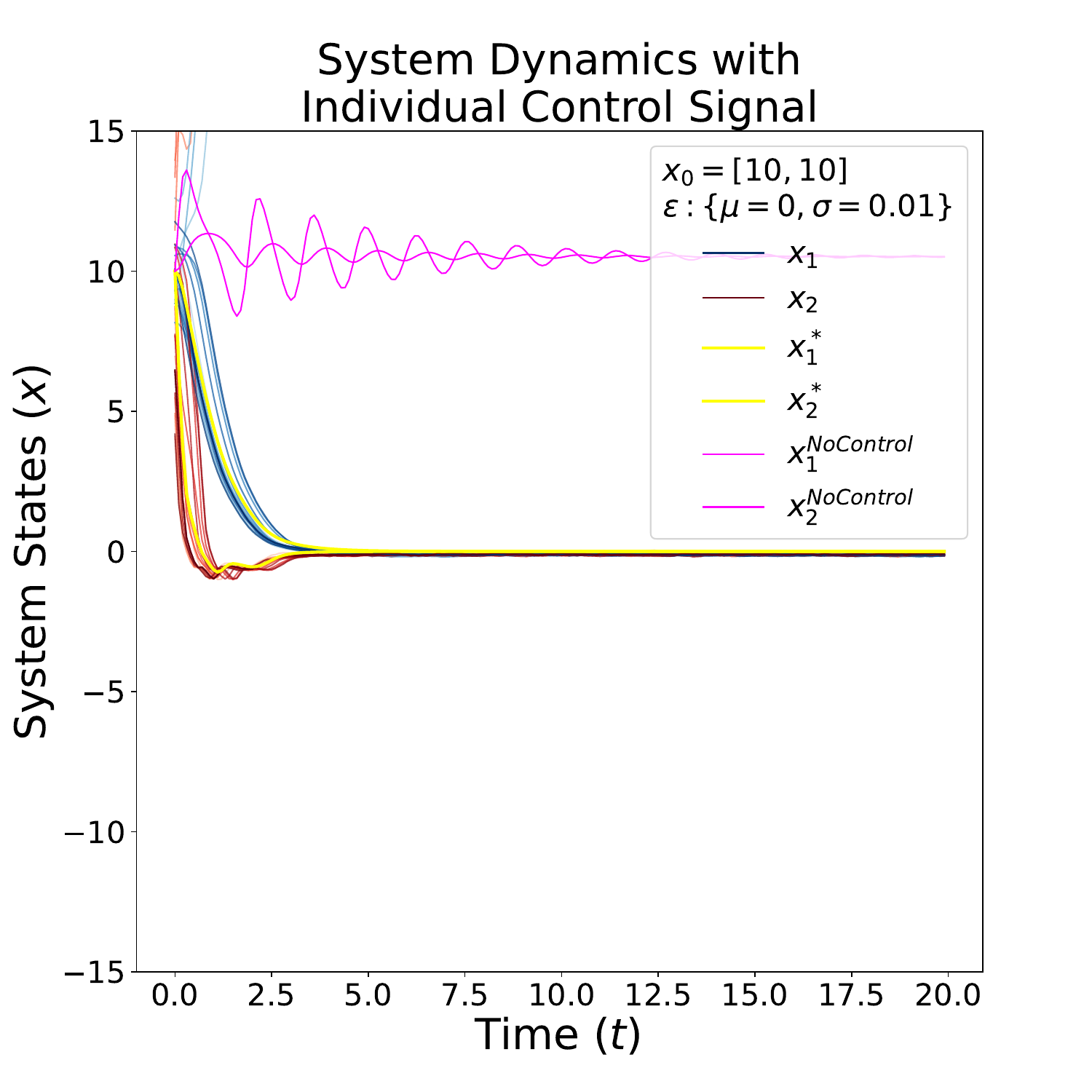}
  \captionof{figure}{\emph{Validation of system dynamics for initial conditions $x_1$, = $x_2$ = 10.0}. \textbf{Panel (a)} shows the system dynamics using the outlier exclusion ensemble control signal policy. \textbf{Panel (b)} shows the system dynamics using the outlier inclusion ensemble control signal policy. \textbf{Panel (c)} shows the system dynamics using individual control signal policies, as well as the system dynamics under zero/no control signal policy.}
  \label{fig:system-dynamics}
\end{center}
\end{figure}

\textbf{Analytical data generation.} The process to generate data for model training and validation purposes is as follows. Define a uniform mesh grid of the relevant state $x$ variables (i.e. a two-dimensional problem would to define $x_1$ and $x_2$), spanning a chosen numerical range (i.e. $-10.0 \ge [x_1, x_2] \le 10.0$) and discretization resolution (i.e. 500 elements). Then, for each grid point $x_i$ in the state space $x_n$ (i.e. $x_{1_i} \in x_{1_m}, x_{2_i} \in x_{2_m}$), the corresponding variables are calculated per \cref{eqn:dierks-system}, stored in separate data structures, and then written to file: the costate $\lambda^*$ is created by evaluating the expression $[x_1, 2x_2]$, the control signal $u^*$ is created by evaluating the expression $\frac{-1}{2} [x_1, 2 x_2] f_2([x_1, x_2])$ where $f_2$ is defined in \cref{eqn:dierks-system}, and the cost-to-go $\mathcal{J}^*$ is created by evaluating the expression $\frac{1}{2} x_1^2 + x_2^2$.

\textbf{Analytical solution to the Hamiltonian.} Leveraging Julia's DifferentialEquations package \citep{rackauckas2017differentialequations}, and the Tsitouras 5/4 Runge-Kutta 
method (free 4th order interpolant) ODE solver (Tsit5) \citep{tsitouras2011runge}, the Hamiltonian to \cref{eqn:dierks-system} is computed analytically by combining a running cost term $\mathcal{J}^*$ and costate contributions $\lambda^*$ with the system's drift $x$ and control dynamics $u^*$. Firstly, \cref{eqn:dierks-system} is created in Julia as an ODE function that defines the closed-loop state derivatives ($\dot{x}$). This ODE function computes the optimal control signal as $u^*=-\frac{1}{2}\,\lambda^T\,f_2(x)$, then sets the value of $\dot{x}$ equal to the evaluated expression of \cref{eqn:affine-system}. This ODE is then sent into the ODEProblem function from Julia's DifferentialEquations package \citep{rackauckas2017differentialequations}. The ODEProblem function specifies everything needed to solve an initial value problem. The initial state $x_0$ of $[x_1=-10.0,x_2=-10.0]$ was used, along with a timespan of $[t_0 = 0.0, t_f=5.0]$. The ODE is then integrated over this timespan using the Tsitouras 5/4 Runge-Kutta 
method (free 4th order interpolant) solver \citep{tsitouras2011runge}, with a relative tolerance of 1e-7 and an absolute tolerance of 1e-6. Subsequently, this time series of solutions is fed into the Hamiltonian function defined as \cref{eqn:hamiltonian-expanded} to be evaluated.

\textbf{Sampling boundary data for model training.} During HJB PINN loss training, boundary data as detailed in \cref{eqn:boundary-data} is randomly sampled (with replacement) according to batch training size (400). This sampling of boundary data is \emph{only} used in the first component of the HJB PINN loss function in \cref{eqn:hjb-pinn-loss-design-ensemble}. The purpose of this is to keep the problem constrained while the network seeks to find the optimal cost-to-go, and therefore the corresponding optimal control signal.

\textbf{Sampling data for model training.} The randomly sampled (with replacement) data used in the second loss component of \cref{eqn:hjb-pinn-loss-design-ensemble} contains a mixture of \emph{both} interior and exterior (boundary) data points. The number of sampled data points is equal to the batch size set during training (100 during warm-start training, and 400 during HJB PINN loss training). This can be thought of as regular data sampling, and it is also the same method used during warm-start training.

\textbf{Network parameter weight conditioning (Warm-start training).} The training process begins with a warm-start phase, where each network in the ensemble is pre-trained on suboptimal cost-to-go data generated as detailed above. This step initializes the network weights to approximate the cost-to-go function, providing a stable starting point for subsequent optimization. Note that the numerical method Tsit5 is used to solve the control problem forward in time, thus provides analytical trajectory information. However, it itself does not minimize the Hamiltonian nor solve the HJB equation. Therefore, the HJB PINN loss is necessary to further refine the network weights.

\textbf{HJB PINN loss training.} Following the warm-start phase, the networks are fine-tuned using a loss function derived from the Hamilton-Jacobi-Bellman (HJB) equation. The loss function is defined in \cref{eqn:hjb-pinn-loss-design-ensemble} and includes two components: a term enforcing the boundary conditions and a term penalizing deviations from the HJB equation. \Cref{eqn:hjb-pinn-loss-design-ensemble} is equivalent to \cref{eqn:HJB-PINN-loss}, but adds ensemble notation for clarity.
\begin{equation}
    \label{eqn:hjb-pinn-loss-design-ensemble}
    \begin{split}
    \mathscr{L}^{\langle j \rangle}(\theta^{\langle j \rangle}) = & \sum_{i} \biggr \Vert \widehat {\mathcal{J}}_{x_{i_b}}^{\langle j \rangle} - \mathcal{J}^*(x_{i_b}) \biggr \Vert^2 +\\
     \alpha\, & \sum_{i} \biggr \Vert x_i^\top Q x_i + u_i^\top R u_i + \nabla_{x_i} \widehat{\mathcal{J}}^{\langle j \rangle} \cdot \big(f_1(x_i, \theta^{\langle j \rangle}) + f_2(x_i, \theta^{\langle j \rangle}) u_i\big) \biggr \Vert^2,
    \end{split}
\end{equation}

Specifically, the loss \eqref{eqn:hjb-pinn-loss-design-ensemble} ensures that the predicted cost-to-go $\widehat{\mathcal{J}}(x)$ satisfies both the dynamics of the control problem and the optimality condition. This stage allows the networks to refine their approximation of the cost-to-go function and converge toward the true optimal solution for the nonlinear system.

We randomly initialize $N=20$ different networks of the same structure. Each network is first equipped with a Mean Squared Error (MSE) loss function, and its weights are conditioned on cost-to-go analytical data. 
After warm-starting the network weights, the MSE loss function is hot swapped with the HJB PINN loss function. Please note that a boundary condition term is added to the HJB PINN loss to keep the problem constrained. See \cref{eqn:boundary-data} for more information.

The process of solving the control problem \cref{eqn:time-invariant-control-problem} involves the following steps:

\begin{enumerate}
    \item Construct $N$ networks as defined in \cref{eqn:NN-def} with inputs $x$, and output $\widehat{\mathcal{J}}$.
    \item For each network $j$, and for $p \in S$ training epochs, and for $k$ training batch $\in O$ total training data, minimize the HJB PINN loss function \eqref{eqn:hjb-pinn-loss-design-ensemble} using the derived optimal costate $\widehat{\lambda}$ via \cref{eqn:lambda}, and the calculated optimal control $\widehat{u}$ via \cref{eqn:control-signal}. 
    \item Update the network parameters using the following optimizer according to the loss function \cref{eqn:hjb-pinn-loss-design-ensemble}:
    \vspace{-0.5em}
    $$\theta_{k+1}^{\langle j \rangle} = \texttt{AdamW}\left(\theta_{k}^{\langle j \rangle},\nabla_{\theta} L({\theta_{k}^{\langle j \rangle}})\right), \quad k = 1, \dots, O.$$
    \item Repeat step 3 until $p=S$ or until network $j$ converges onto the learning task solution.
    \item Repeat step 2 for network $ j + 1$.
\end{enumerate}
The main technical challenge is learning the optimal cost-to-go in a stable fashion without the loss exploding to infinity. Adding a boundary condition term to the loss function helps constrain the problem. Moreover, the solution to the time-invariant affine HJB PDE equation is not unique, which results in a loss landscape with many local minimas. Warm-starting the network weights on sub-optimal data helps the network not start at a local minima within the manifold of the loss landscape that it cannot escape. Validation of the solution found by the model is discussed in \cref{sec:Results}.

\subsection{Experiment Setup} 

We take N=20 models to comprise the ensemble. All models use the same PyTorch-implemented architecture, and the same weight-conditioned parameters via the warm-starting process. The input layer takes in the system states $x_1$ and $x_2$, which have a dimension of (n, 2, 1), where n is the batch size for training. There are two hidden layers with ten nodes per layer. The output layer is the predicted cost-to-go ($\widehat{\mathcal{J}}$), and has a dimension of (n, 1, 1), where n is the batch size for training. The input and hidden layers use the Tanh activation function, whereas the output layer is linear.

\subsubsection{Warm-started network} 

A model with the same architecture as described above is trained on sub-optimal cost-to-go analytical data generated offline.

This model is used for loading in warm-started network weights into the ensemble of models, that continue training using the HJB PINN loss as detailed in \cref{eqn:hjb-pinn-loss-design-ensemble}. The loss function used at this stage is the Mean Squared Error (MSE) of the predicted cost-to-go and analytical cost-to-go, in training batch sizes of 100. The AdamW optimizer is used with an initial learning rate of 1e-2, an epsilon ($\epsilon$) value of 1e-4, and a weight decay of 1e-2. This optimizer is particularly relevant here, as Adam can be interpreted as an IMEX time-stepping of an underlying momentum ODE, which aligns well with physics-coupled learning \cite{bhattacharjee2024improving}. The learning rate is reduced by a factor of 1e-1 each time the MSE loss reaches a plateaued value after 2 epochs. This warm-started model is trained for 100 epochs.

\subsubsection{Ensemble of HJB PINN networks} Twenty models are initialized by loading in the weights from the warm-started model. These twenty models comprise the ensemble that is used for offline closed-loop control, as shown in \cref{fig:system-dynamics}. The weights of these networks are further refined to reach optimal cost-to-go predictions, and therefore optimal control signals, by training the ensemble of networks using the HJB PINN loss function shown in \cref{eqn:hjb-pinn-loss-design-ensemble}. In this stage, the optimizer used is also AdamW, but the epsilon value is 1e-8, and the weight decay is 1e-7. Additionally, a cyclic learning rate scheduler is used instead of a reduce-on-plateau scheduler. The cyclic learning rate ranges from 1e-5 to 1e-4 with a step size of 20 between the lower and upper bounds. Each model in the ensemble is trained individually for 60 epochs on randomly shuffled batch data, which ensures some variability in their training schemes. At each training step, the model takes two forward passes. Firstly, the model makes predictions on a batch of randomly shuffled data, containing both interior and boundary points. These predictions are used to calculate the HJB component of the loss function shown in \cref{eqn:hjb-pinn-loss-design-ensemble}. Secondly, the model makes predictions on a batch of randomly shuffled boundary only data. These predictions are used to calculate the boundary loss component of the loss function shown in \cref{eqn:hjb-pinn-loss-design-ensemble}.

\subsubsection{Hyper-parameters} Various network configurations were investigated over a sustained time period. Activation functions such as ReLU were tried, but do not provide smooth gradients necessary for automatic differentiation. A Tanh activation does provide smooth gradients. A hidden layer size of up to 10 was explored. However, Tanh runs into vanishing gradient issues with more than three nonlinear activations. Therefore, two hidden layers was used. Nodes per layer of up to 100 were looked into. However, when a model is wider than it is taller, it has a tendency to simply memorize the data, rather than learn from it. In statistics, this is called the bias-variance trade-off. In machine learning, it is known as over-fitting. Therefore, only 10 nodes per hidden layer were used. During warm starting training, AdamW's epsilon value of 1e-4 provides more numerical stability, and its aggressive weight decay value of 1e-2 provides regularization to prevent over-fitting. In contrast, during HJB PINN training, less numerical stability is needed and hence a value of 1e-8 is used. It is important to note that this numerical stability is in the context of conducting AdamW's algorithm, and is not the same as stability terminology in traditional numerical methods when solving the HJB equation. A weight decay value of 1e-8 is necessary during HJB PINN training to give the weights time to converge upon the optimal cost-to-go. A reduce-learning-rate-on-plateau scheduler enables the warm start training to converge upon the sub-optimal mappings between system states and their corresponding cost-to-go, yet still gives the network the ability to make minute adjustments when in a plateau region of the objective function manifold. In contrast, a cyclic learning rate enables the HJB PINN training to traverse multiple local minima when attempting to converge upon the optimal mapping between system states and their corresponding cost-to-go. A cyclic learning rate range of 1e-8 to 1e-1 was investigated, and in the end the range of 1e-5 to 1e-4 was found to perform the best.

\subsection{Ensemble Control PINN Results}
\label{sec:Results}

The experiment setup and process of training to solve the above system's optimal cost-to-go, and subsequently the corresponding optimal control signal, through the HJB equation is explained in \cref{sec:Experiment} and \cref{eqn:hjb-pinn-loss-design-ensemble}, respectively. \Cref{fig:system-dynamics} shows the offline closed-loop control of the nonlinear system in \cref{eqn:dierks-system} with three different control policies. The system is sent probing control signals from the controller along a time-span range of $\{t_0 = 0.0, t_f = 20.0\}$, with a time-step of $t_{step} = 0.01$. The initial conditions of $x_1$ and $x_2$ at $t_{0}=0.0$ are perturbed randomly for each model in the ensemble. Moreover, at each time step along the full time-span, $x_1$ and $x_2$ are also perturbed randomly. This noise is added to the system state variables in the form of a random normal distribution with mean $\mu = 0.0$ and standard deviation $\sigma = 0.01$. The noise level can be adjusted, but increasing the perturbation too much decreases the chances of the controllers (ensemble and individual) from keeping the system state on the path of an admissible trajectory. This is because with too much variance in starting conditions, the system may become uncontrollable. Indeed, some systems fail to meet the admissible trajectory, and their paths go off the plots' y-axis range in \cref{fig:system-dynamics}. That being said, it is noteworthy that the three different control policies can stabilize the system through dynamical perturbation on-the-average, especially considering that the models were not trained on perturbed data themselves. The three control policies demonstrated in \cref{fig:system-dynamics} are detailed below.

\begin{figure}
\begin{center}
  (a) \includegraphics[width=0.44\textwidth, trim=4 4 4 4,clip, keepaspectratio]{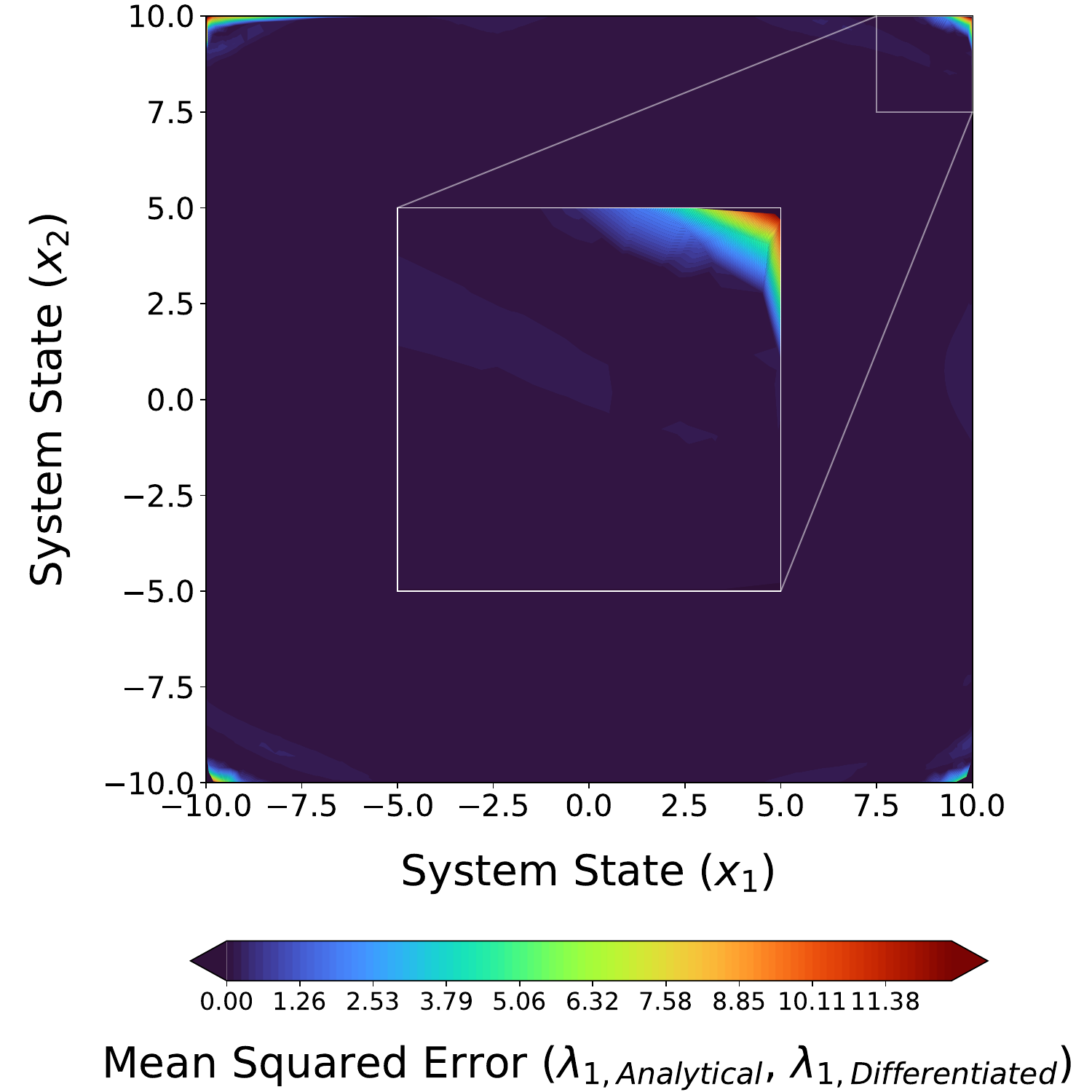}
  (b) \includegraphics[width=0.44\textwidth, trim=4 4 4 4,clip, keepaspectratio]{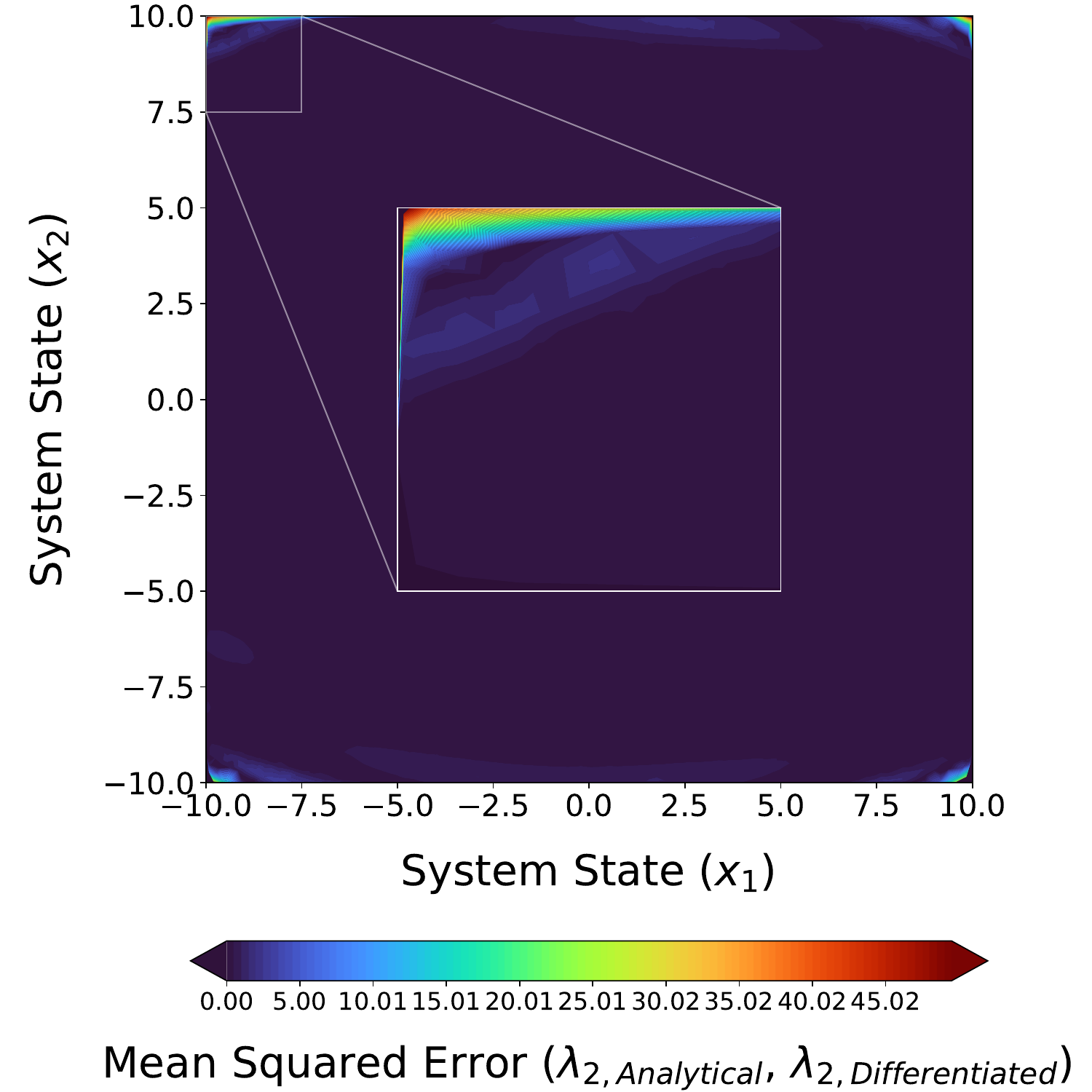} \\
  (c)  \includegraphics[width=0.44\textwidth, trim=4 4 4 4,clip, keepaspectratio]{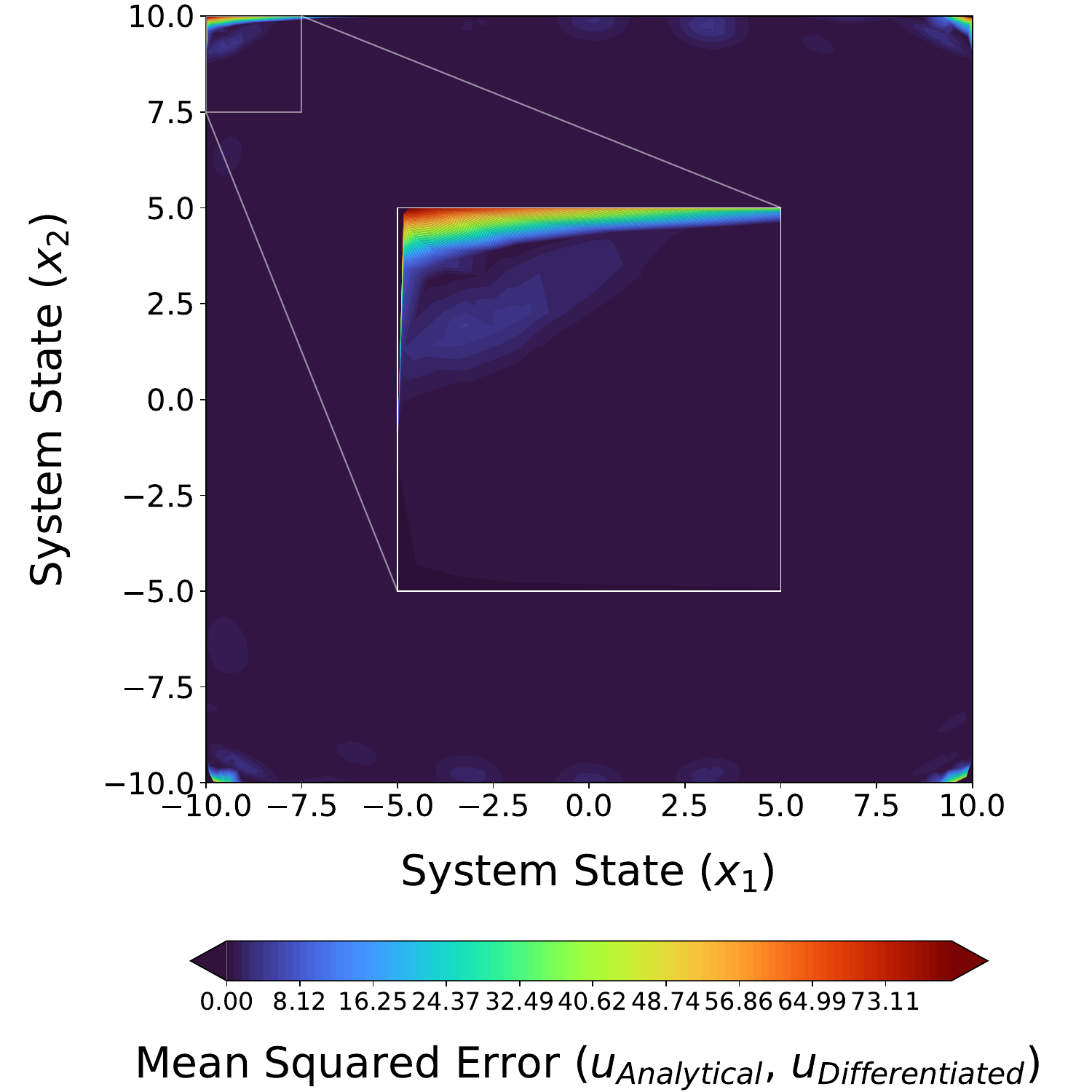}
  \captionof{figure}{\emph{Mean Squared Error (MSE) plots for the derived Lambda ($\hat\lambda^e$) and the calculated optimal control signal ($\hat u^e$).} Panels (a) and (b) show the MSE between the analytical Lambda ($\lambda^*$) and the differentiated Lambda ($\hat\lambda^e$) from the learned Cost-to-Go ($\widehat{\mathcal{J}}^e$) with respect to the system state ($x$), for a given model among the ensemble. \textbf{Panel (a)} shows the MSE($\lambda_1^*, \hat \lambda_1^e$). \textbf{Panel (b)} shows the MSE($\lambda_2^*, \hat \lambda_2^e$). \textbf{Panel (c)} shows the MSE between the analytical control signal ($u^*$) and the differentiated optimal control signal ($\hat u^e$), for a given model in the ensemble. $\hat u^e$ is calculated via the differentiated Lambda ($\hat \lambda^e$) from the learned Cost-to-Go ($\widehat{\mathcal{J}}^e$) with respect to the system state ($x$).}
  \label{fig:mse-lambda}
\end{center}
\end{figure}

Let $\bar \lambda$ and $\bar u$ represent the ensemble mean of the adjoint and the ensemble mean of the control signal, respectively. Likewise, let $\hat \lambda^e$ represent the adjoint for a given individual control $\hat u^e$. The mean squared error (MSE) for the derived adjoint ($\hat\lambda^e$) and corresponding optimal control ($\hat u^e$) for one of the given ensemble models are shown in \cref{fig:mse-lambda}. The vast majority of the errors are effectively zero. The largest errors occur at the boundary conditions, and a zoomed-in view on those regions are provided in \cref{fig:mse-lambda} for convenience.

\begin{figure}
\begin{center}
  (a) \includegraphics[width=0.4\textwidth]{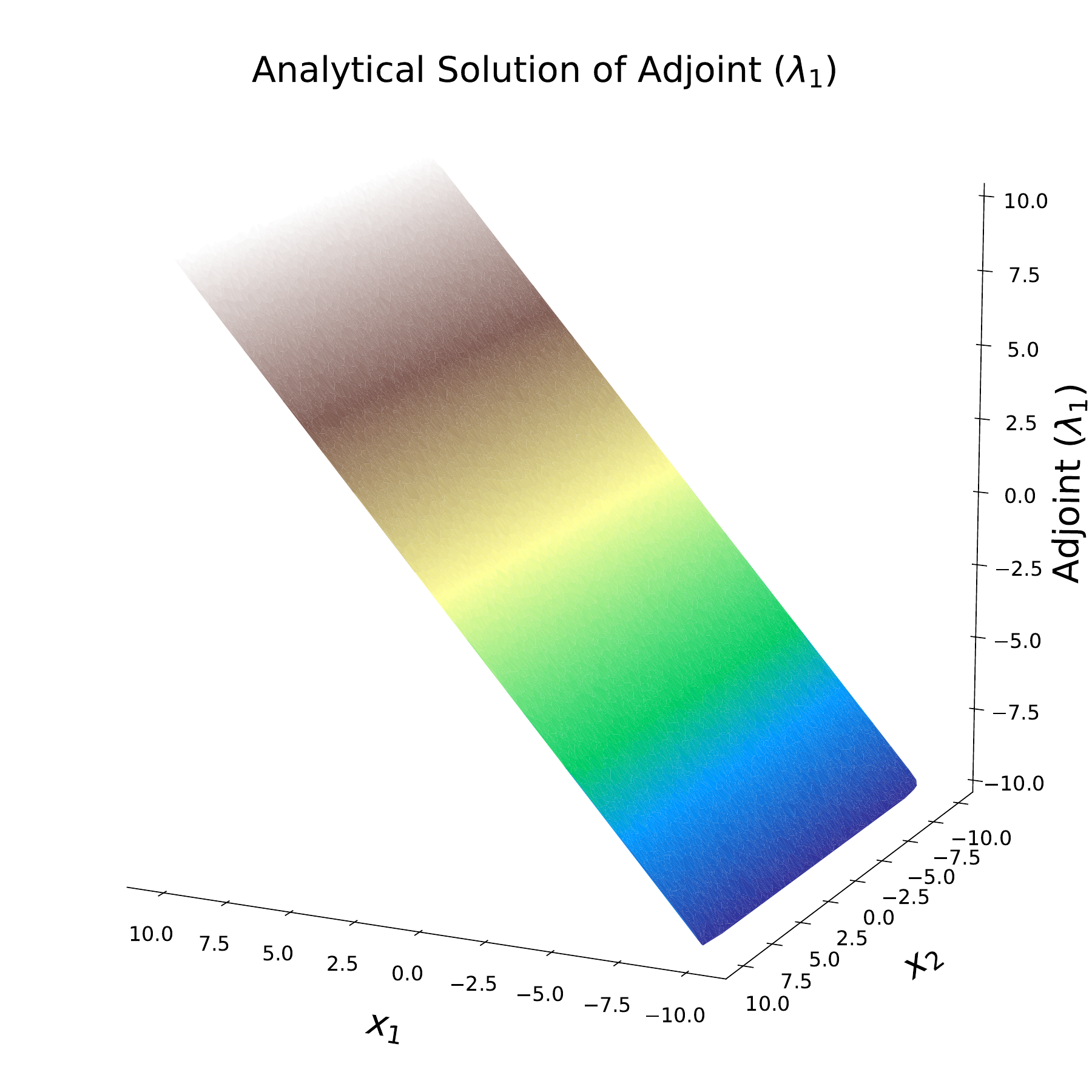}
  (b) \includegraphics[width=0.4\textwidth]{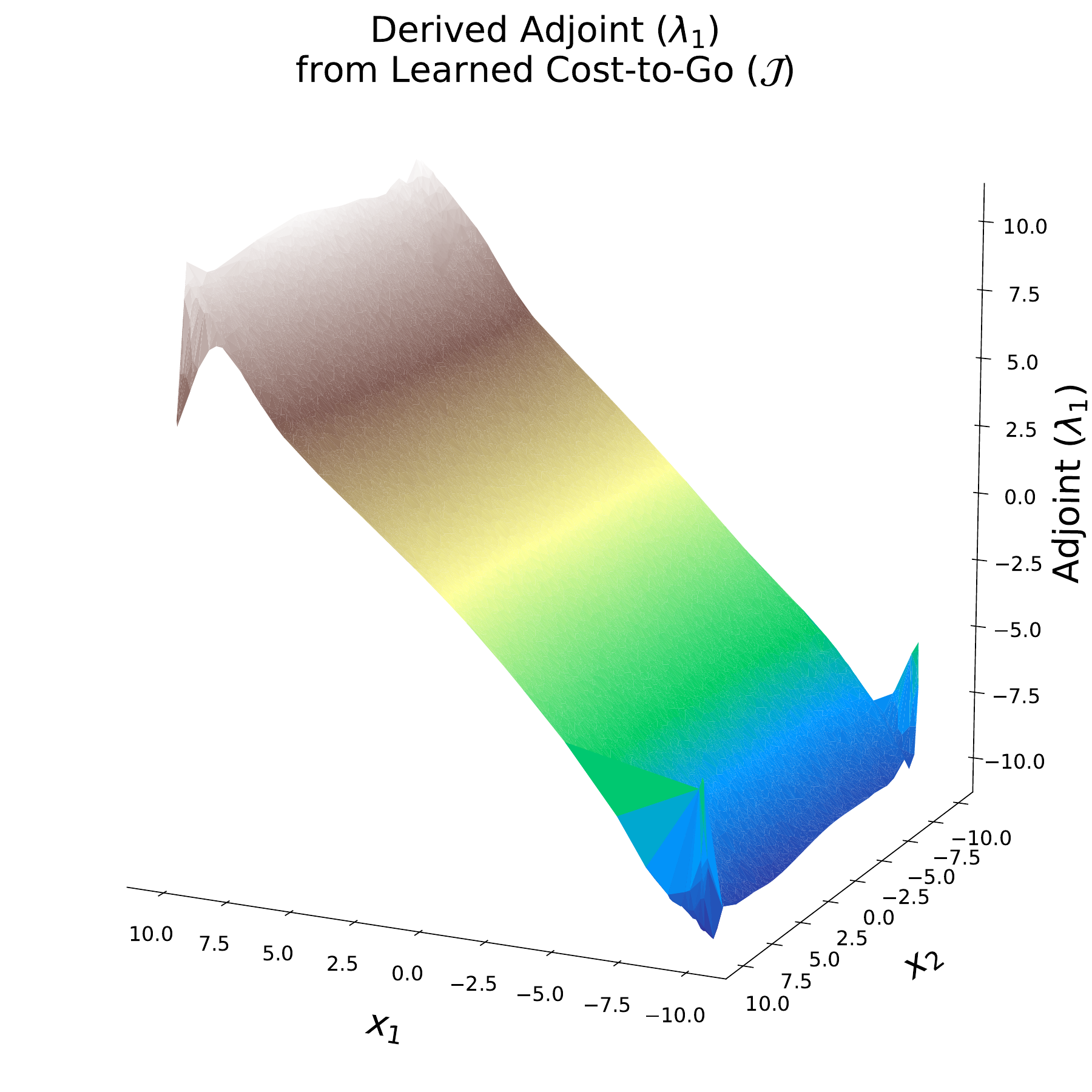}\\
  (c) \includegraphics[width=0.4\textwidth]{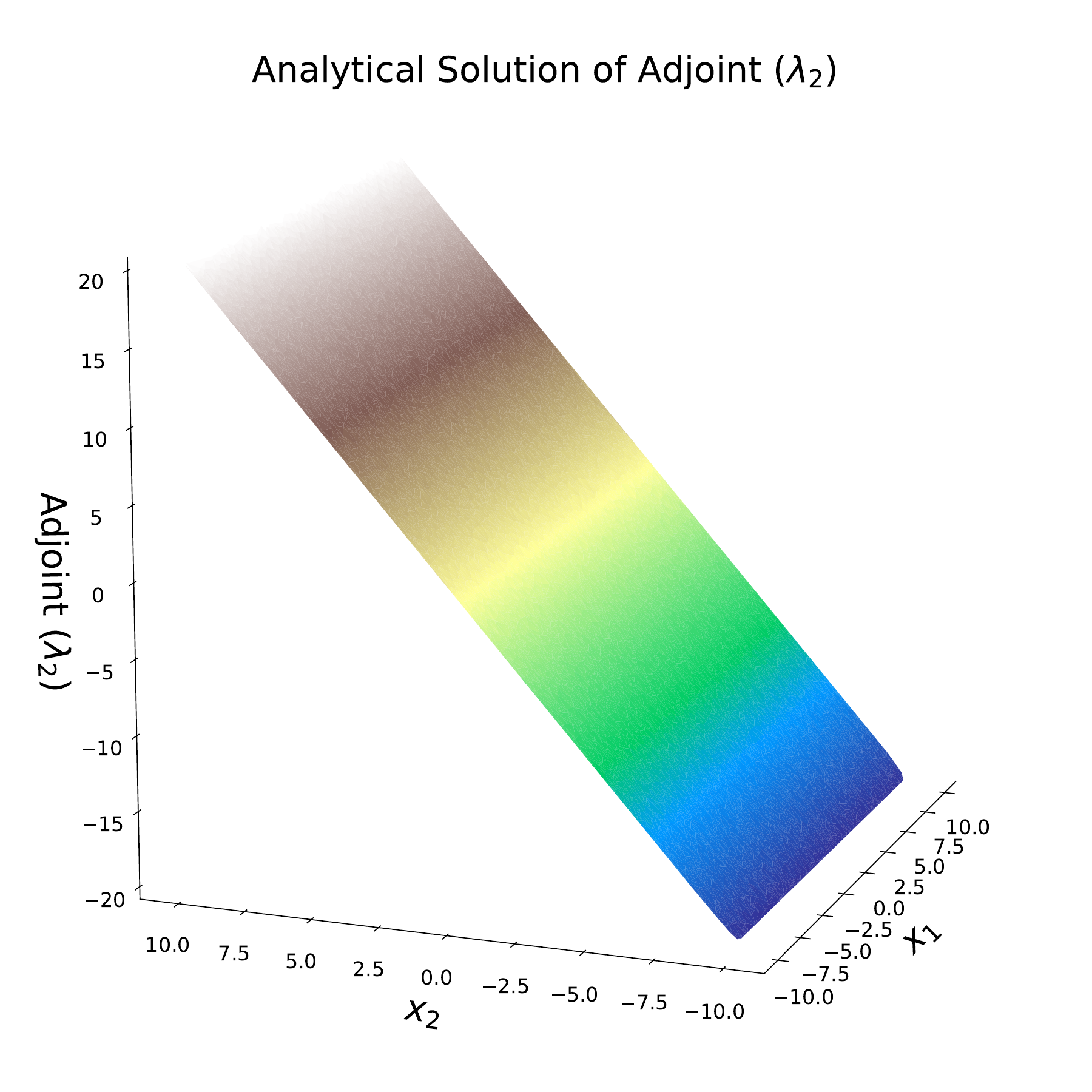}
  (d) \includegraphics[width=0.4\textwidth]{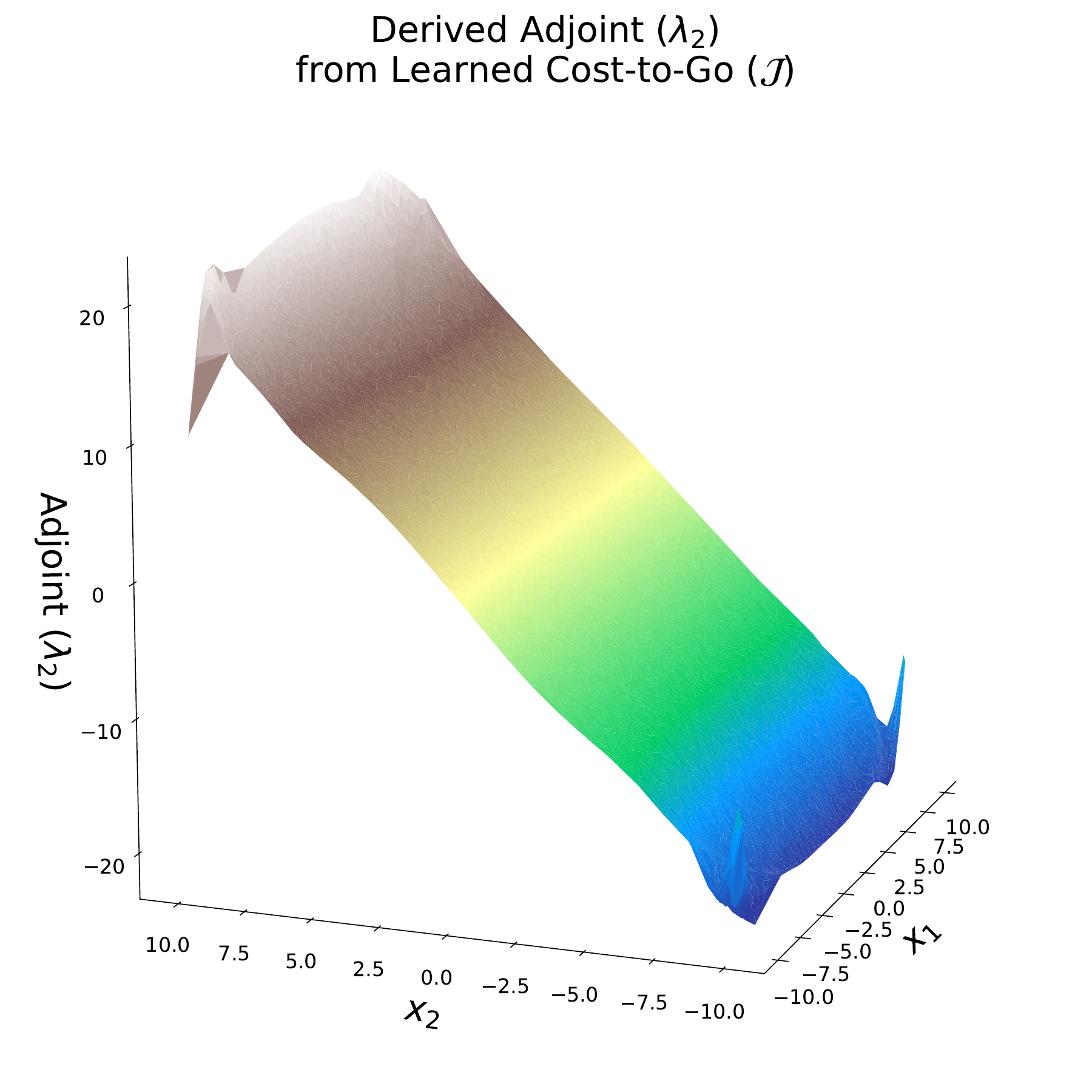}\\
  (e) \includegraphics[width=0.4\textwidth]{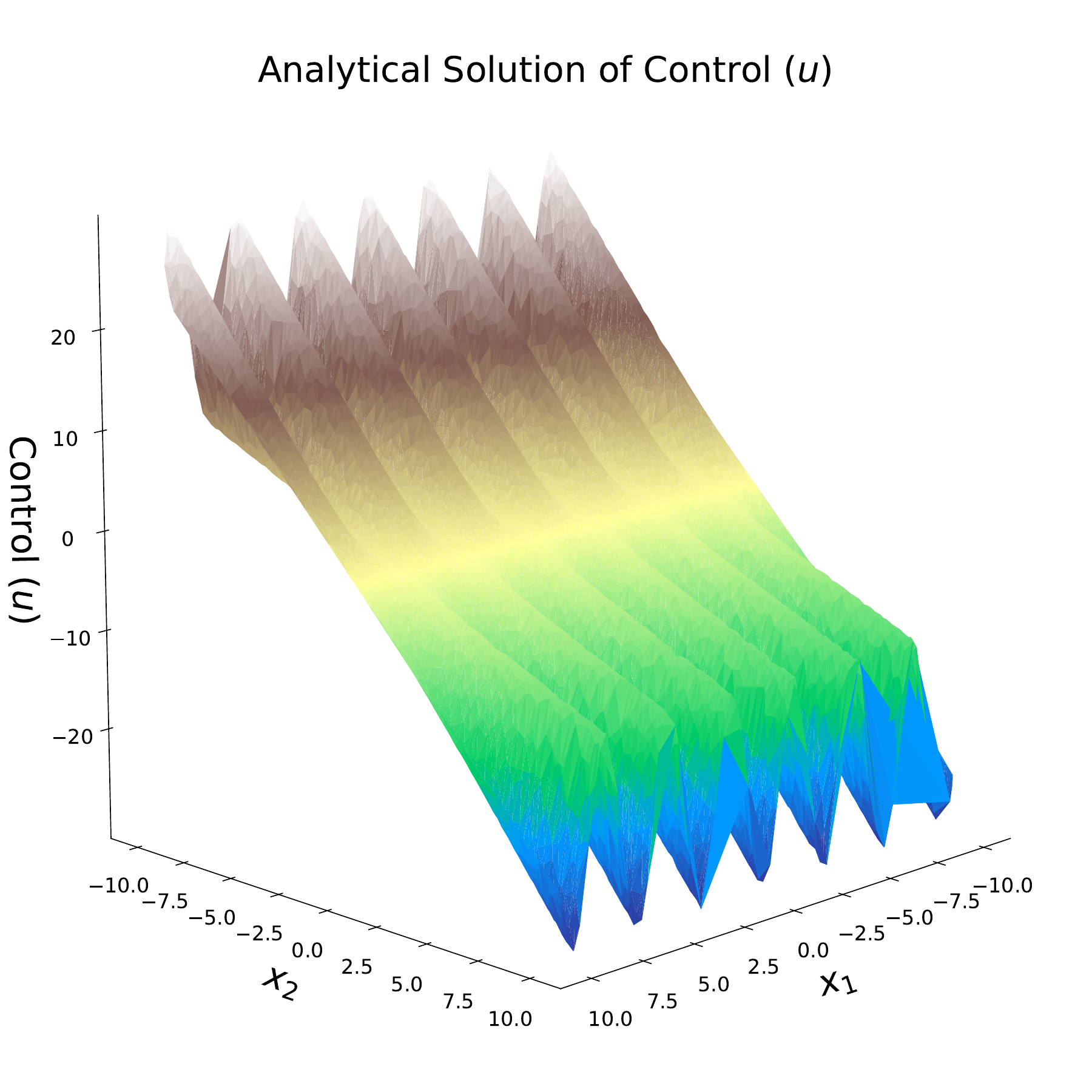}
  (f) \includegraphics[width=0.4\textwidth]{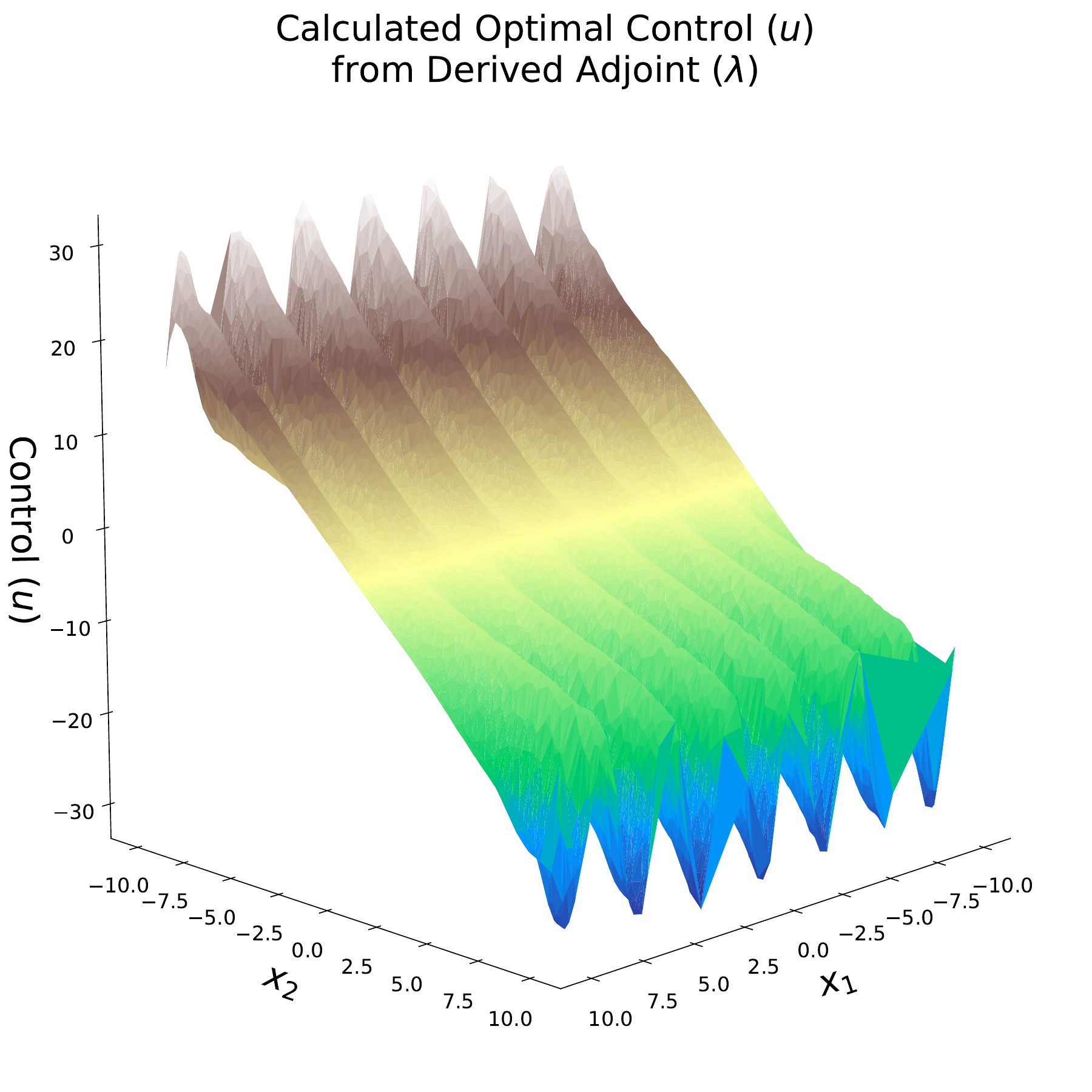}\\
  \captionof{figure}{\emph{Three-dimensional reconstruction and comparison between the reference analytical solution and the learned network solution for both the adjoint ($\lambda_1$ and $\lambda_2$) and control ($u$).} \textbf{Panel (a)} shows the analytical solution $\lambda_1^*$, whereas \textbf{panel (b)} shows the learned solution of $\hat \lambda_1^e$. \textbf{Panel (c)} shows the analytical solution of $\lambda_2^*$, whereas \textbf{panel (d)} shows the learned solution of $\hat \lambda_2^e$. \textbf{Panel (e)} shows the analytical solution of $u^*$, whereas \textbf{panel (f)} shows the learned solution of $\hat u^e$.
  }
  \label{fig:3D-lambda-control}
\end{center}
\end{figure}

\begin{wrapfigure}{R}{6cm}
\centering
\includegraphics[width=6cm]{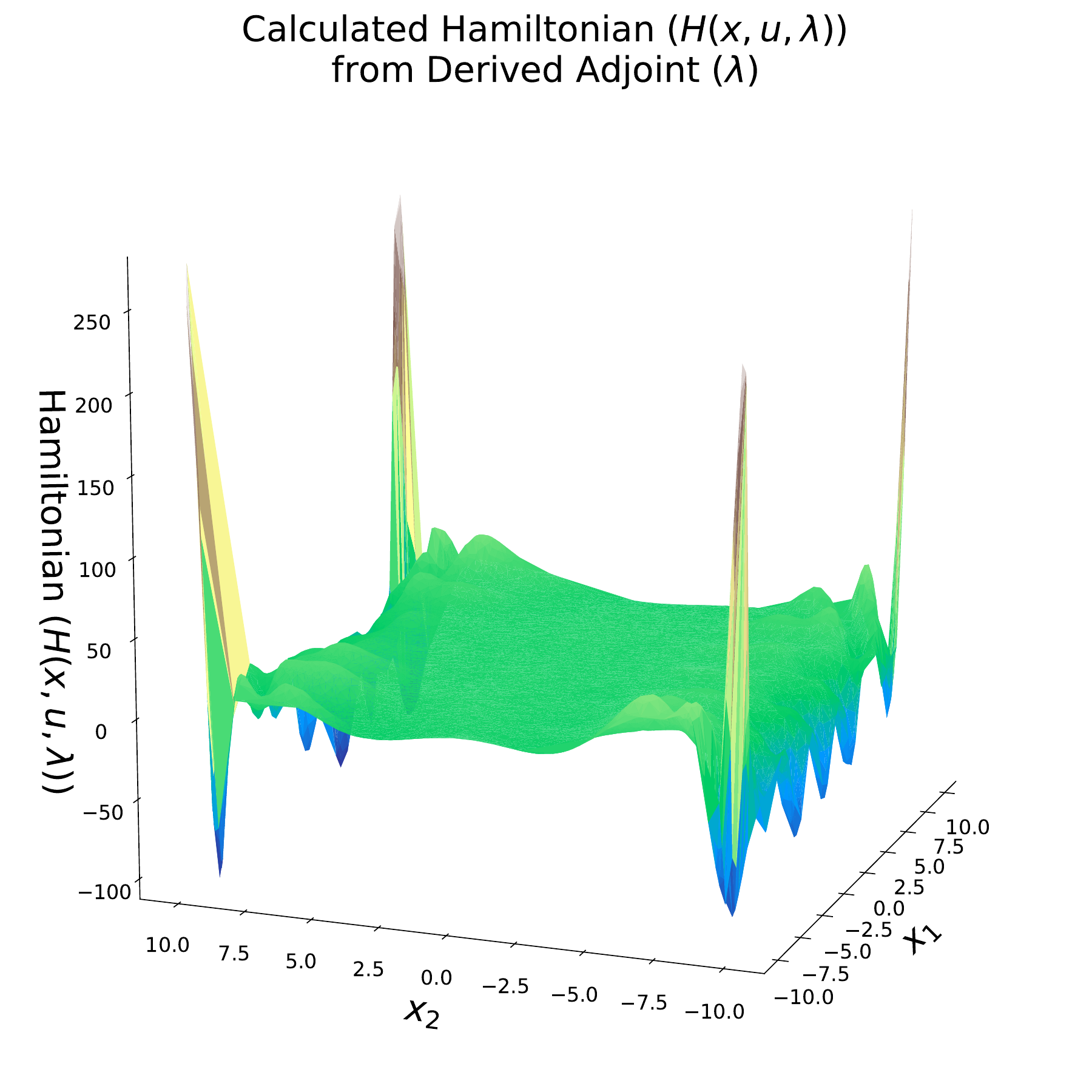}
  \captionof{figure}{\emph{Three-dimensional reconstruction of the Hamiltonian from the network's derived adjoint ($\hat{\lambda}^e$) and calculated optimal control ($\hat{u}^e$) }. The reference solution of the Hamiltonian is $H^*(x, u, \lambda) = 0$.}
  \label{fig:3D-hamiltonian}
\end{wrapfigure}

\Cref{fig:3D-lambda-control} is a three-dimensional reconstruction of both the reference solution and a given network's learned solution. Specifically, in the first and second panel rows, a comparison of the reference adjoint ($\lambda_1^*$ and $\lambda_2^*$) and the learned adjoint ($\hat \lambda_1^e$ and $\hat \lambda_2^e$) are shown. These are the same adjoints ($\hat \lambda^e$) as in \cref{fig:mse-lambda}. Similarly, the third row of panels in \cref{fig:3D-lambda-control} shows the reference solution of the control signal ($u^*$) to that of the network control signal ($\hat u^e$). The smooth planes of the analytical solution are well matched with those of the learned solution, albeit with slight disturbances at the boundary conditions, which are represented in the zoomed-in region of the MSE plots in \cref{fig:mse-lambda}. Additionally, a three-dimensional reconstruction of the corresponding learned Hamiltonian is shown in \cref{fig:3D-hamiltonian}.

\textbf{Outlier including ensemble mean control signal.} In this control policy, the mean control signal is applied to each of the twenty systems, regardless of the presence  of outlying control signals. The mean control signal is calculated by taking the average signal strength based on each of the twenty individual models that make up the ensemble. This control policy is included for posterity, and is not the most adaptable, since by definition it would include control signals corresponding to systems that are beyond controllable and hence negatively influence the rest of the ensemble. Still, it does a similar job as compared to the outlier excluding control policy.

\textbf{Outlier excluding ensemble mean control signal.} For this control policy, the Chauvenet criterion (at 95\% confidence level) is used to identify outlying control signals among the ensemble. Control signals that meet the Chauvenet criterion are excluded from the mean control signal calculation. However, in the edge-case that all control signals in the ensemble are ``thrown out'' at the current time-step, the control policy reverts to applying individual control signals to each respective dynamical system. This is done with the hope of bringing outlying systems closer back to an admissible trajectory, such that the mean control signal of the entire ensemble can continue to be applied at future time-steps.

\textbf{Individual control signal.} This control policy is straight forward, and offers a non-ensemble approach. At each time-step, the control signal for the respective perturbed dynamical system is applied. A mean control signal is not always applicable, or effective for stabilizing a system. This is best seen in the middle row of \cref{fig:system-dynamics}, where the ensemble control policies center around, but not on, the non-perturbed analytical reference solution with a starting initial condition of $x_1, x_2 = 0.0$. In contrast, the individual control signal policy is still able to stabilize the system to zero through noisy data.

\section{Conclusions}
\label{sec:Conclusions}

This work proposes an ensemble PINN control methodology based on solving the HJB equation on an infinite time horizon. The novel aspects of this work include an outlier excluding ensemble mean control signal policy, the non-reliance of the averaging principle from multi-agent systems, and solving of the HJB equation for the infinite time horizon case optimally without relying on sufficient approximation via finite horizon.

Success of the proposed methodology is demonstrated via a two-state continuous nonlinear affine system. Three control policies are put forth in this work: (1) ensemble control with outlier inclusion, (2) ensemble control with outlier exclusion, and (3) individual member control.

Validation of results are shown by plotting the system dynamics in \cref{fig:system-dynamics}, which illustrate that the network is able to learn the optimal control signal matching that of the analytical solution. Consequently, the network is able to stabilize the system's dynamics to zero with probing control signals even in the presence of data perturbation of the system state variables. Moreover, additional system dynamics with varying initial conditions are put forth in \cref{appendix:expanded-results}. Further quantitative validation includes plotting the Mean Squared Error (MSE) of the adjoint ($\lambda$) and the control ($u$) in \cref{fig:mse-lambda}. Additionally, \cref{fig:3D-lambda-control} presents a three-dimensional reconstruction of the both the analytical adjoint ($\lambda^*$) and control ($u^*$) and compares them to those of a given network's learned adjoint ($\hat \lambda ^e$) and control ($\hat u ^e$).

Future work possibilities include extending this framework to more complex problems such as cart pole and inverted pendulum. The open source Python library called Gymnasium offers a good environment for testing frameworks on classic control problems \citep{towers2024gymnasium}. In addition, a sensitivity study would shed light on the extent to which the three control policies present in this work can handle increasing amounts of noise in the data.












\section{Acknowledgments}
This research is supported by NSF awards DMS--2411069, DMS--2436357, and the Computational Science Laboratory at Virginia Tech. The work of Arash Sarshar uses Jetstream2 at Indiana University through allocation CIS230277 from the Advanced Cyberinfrastructure Coordination Ecosystem: Services \& Support (ACCESS) program, which is supported by National Science Foundation grants \#2138259, \#2138286, \#2138307, \#2137603, and \#2138296.

\section{Declaration of competing interest}

The authors declare that they have no known competing financial interests or personal relationships that could have appeared to influence the work reported in this paper.

\section{Declaration of generative AI and AI-assisted technologies in the writing process}

During the preparation of this work the authors used OpenAI's ChatGPT in order to assist with crafting the formal notation of the presented methodologies and their underlying concepts. After using this tool/service, the authors reviewed and edited the content as needed and take full responsibility for the content of the published article.

\section{CRediT authorship contribution statement}

\textbf{Jostein Barry-Straume:} Formal analysis, Investigation, Methodology, Software, Validation, Writing---original draft, Writing---review \& editing. 
\textbf{Adwait D. Verulkar:} Data curation, Methodology, Software, Writing---original draft, Validation.
\textbf{Arash Sarshar:} Conceptualization, Formal analysis, Methodology, Supervision, Writing---review \& editing.
\textbf{Andrey A. Popov:} Conceptualization, Methodology, Supervision, Writing---review \& editing.
\textbf{Adrian Sandu:} Conceptualization, Funding acquisition, Methodology, Project administration, Supervision, Writing---review \& editing.

\appendix

\section{Expanded Results}
\label{appendix:expanded-results}

The figures in this section offer a view into expanded results for greater visual clarity. More importantly, the figures show that: (1) without a control signal the system cannot stabilize to zero unless its starting conditions are already zero, and (2) the neural network is able to learn the optimal control signal solution matching that of the analytical control.

There are $N=20$ trained models that make up the ensemble. For the figures shown in this section, the blue lines show the system dynamics for $x_1$ corresponding to each model in the ensemble. Likewise, the blue lines represent the system dynamics for $x_2$. 
The yellow lines show the system dynamics ($x_1$ and $x_2$) under direction of the analytical control signal with zero perturbation in the data.
Similarly, the magenta lines represent the system dynamics ($x_1$ and $x_2$) with no control signal probing the system.
At each time step, a random perturbation is added to the system states $x_1$ and $x_2$ in the form of a random normal distribution of mean zero and standard deviation of 0.01.

\begin{figure}[H]
    \centering
    \begin{subfigure}[b]{0.44\textwidth}
        \centering
        \includegraphics[width=\linewidth]{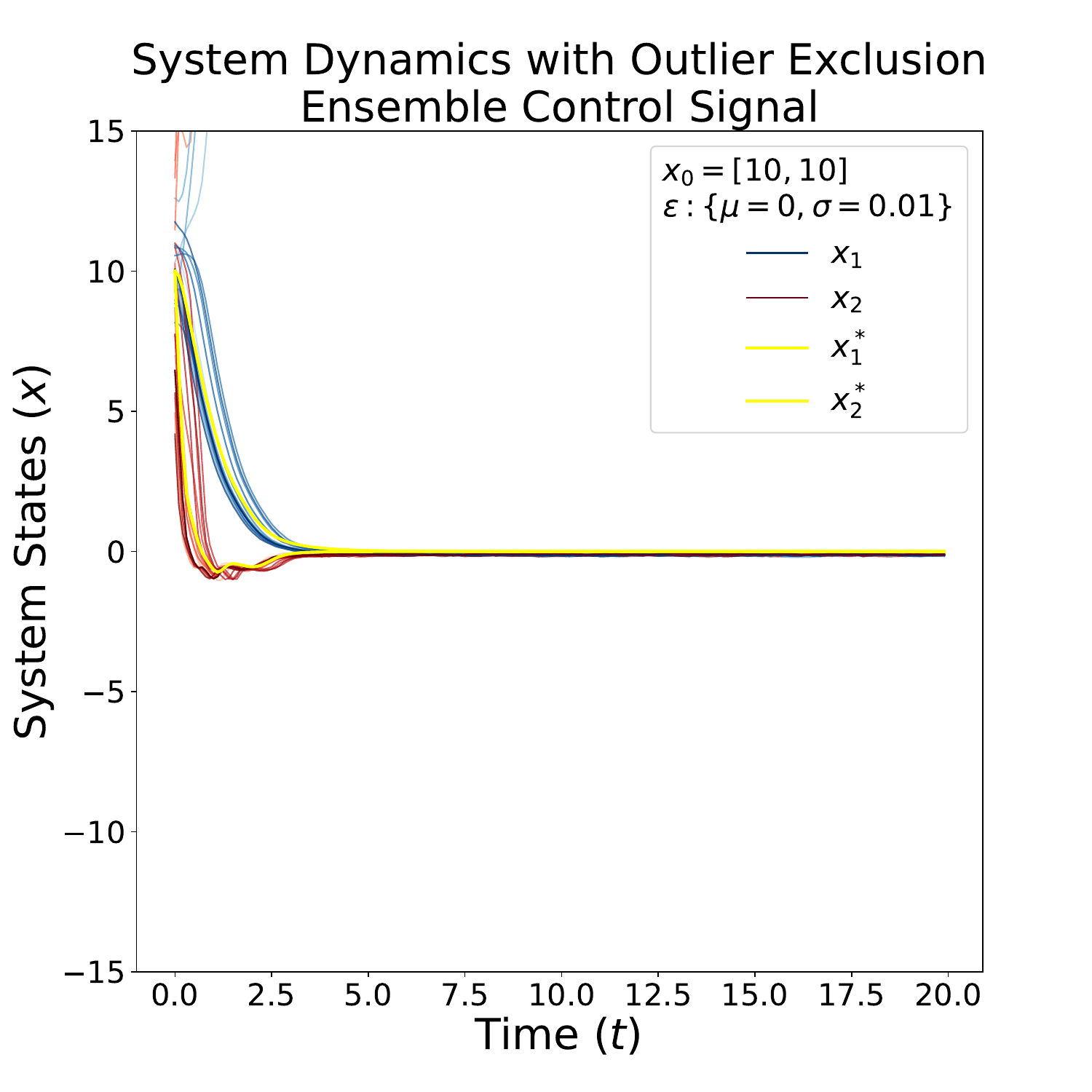}
        \caption{Validation of the system dynamics, with initial conditions $x_1$, $x_2$ = 10.0, resulting from applying a mean control signal ($\bar u$) that excludes outlying systems.}
    \end{subfigure}
    \hfill
    \begin{subfigure}[b]{0.44\textwidth}
        \centering
        \includegraphics[width=\linewidth]{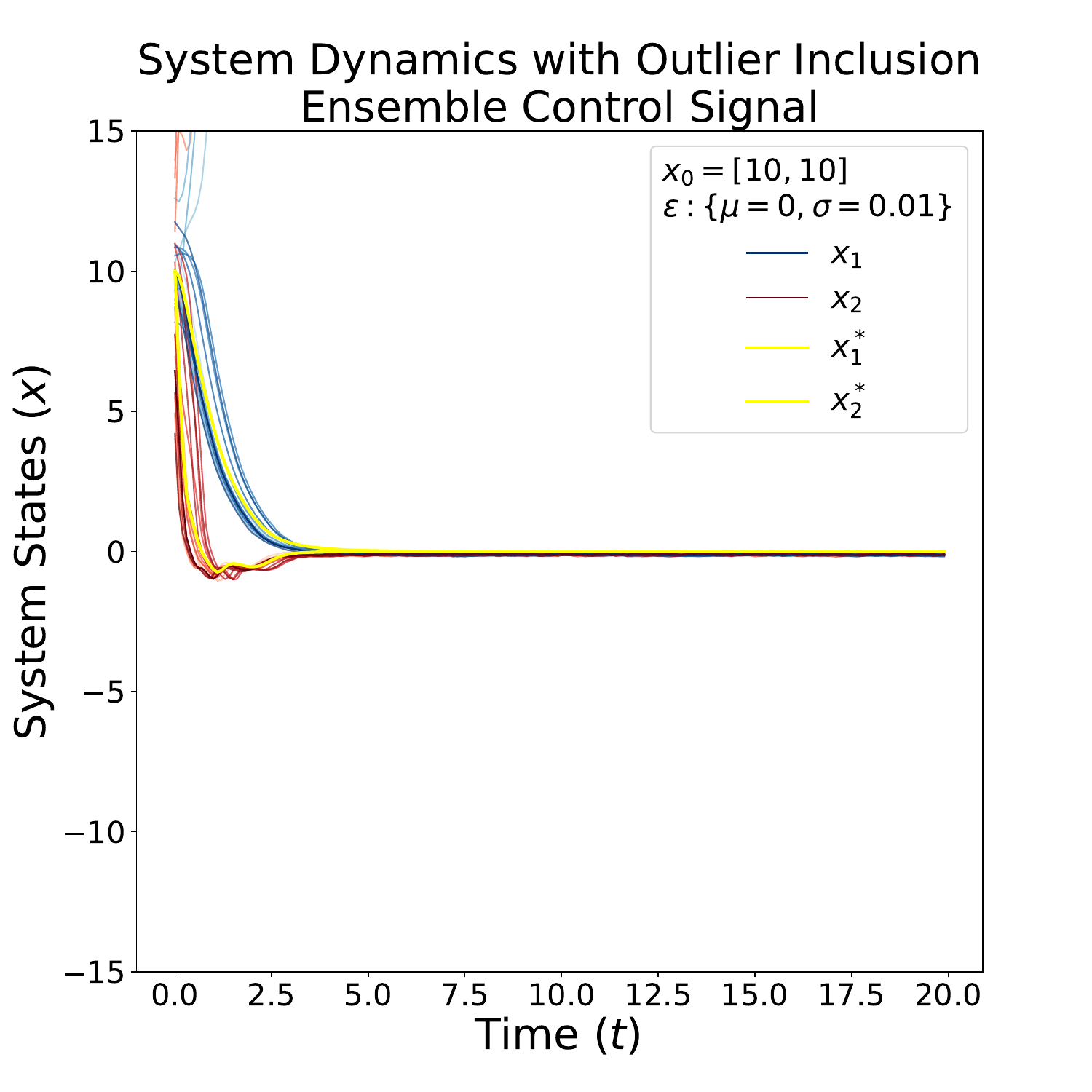}
        \caption{Validation of the system dynamics, with initial conditions $x_1$, $x_2$ = 10.0, resulting from applying a mean control signal ($\bar u$) that includes outlying systems.}
    \end{subfigure}\\
    \vspace{1em}
    \begin{subfigure}[b]{0.44\textwidth}
        \centering
        \includegraphics[width=\linewidth]{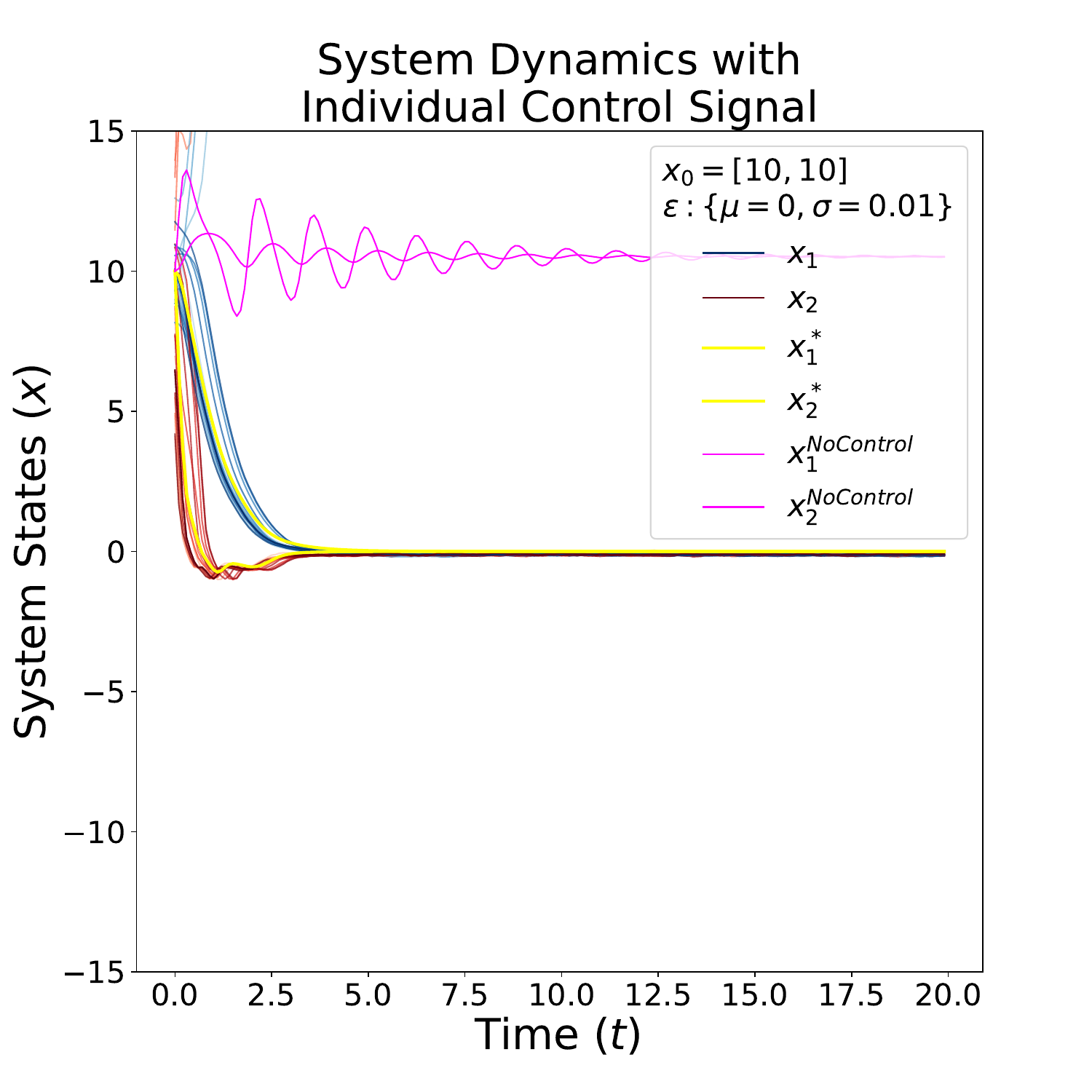}
        \caption{Validation of the system dynamics, with initial conditions $x_1$, $x_2$ = 10.0, resulting from applying each individual model's control signal ($\hat u^e$) to their respective system.}
    \end{subfigure}
\end{figure}

\begin{figure}[H]
    \ContinuedFloat
    \centering
    \begin{subfigure}[b]{0.44\textwidth}
        \centering
        \includegraphics[width=\linewidth]{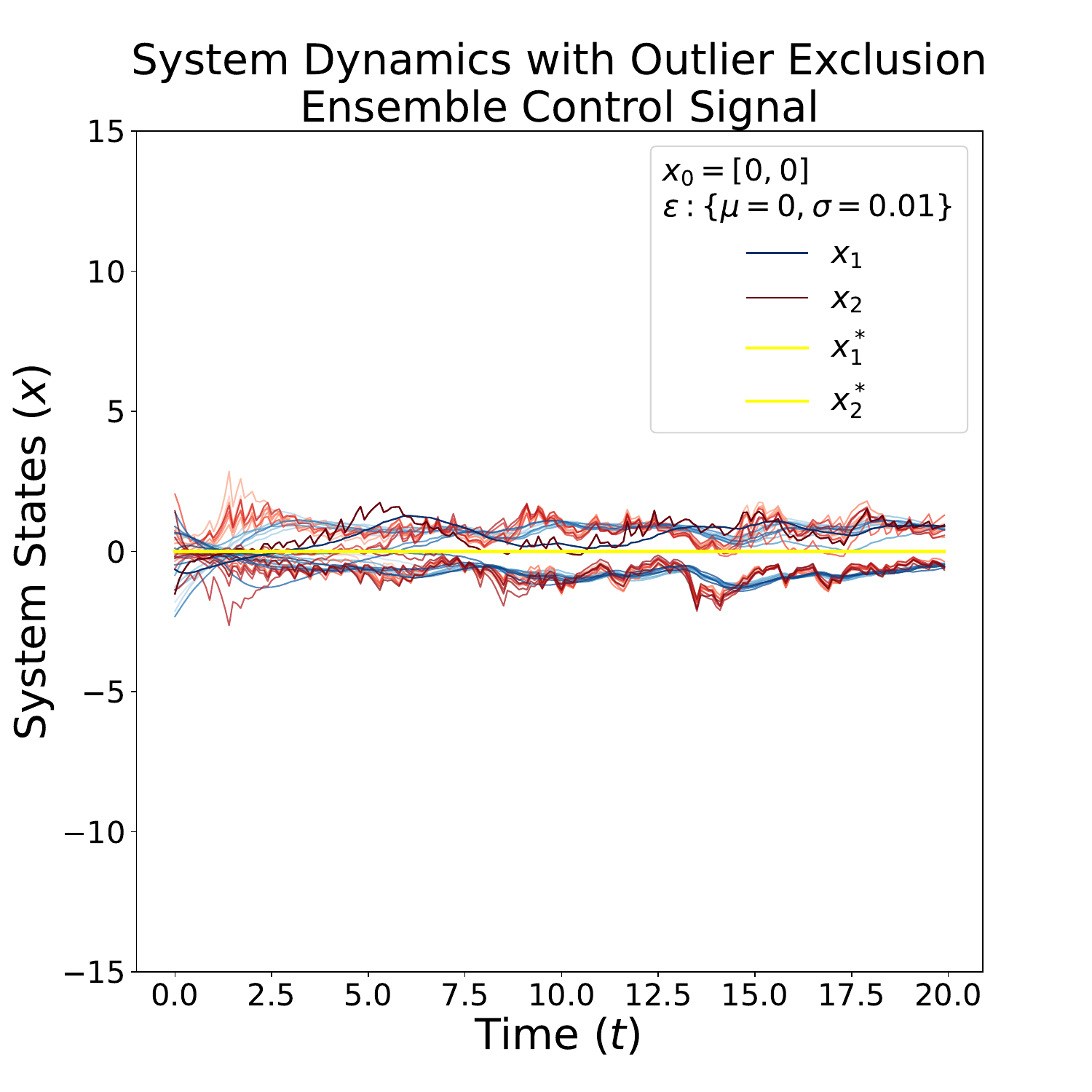}
        \caption{Validation of the system dynamics, with initial conditions $x_1$, $x_2$ = 0.0, resulting from applying a mean control signal ($\bar u$) that excludes outlying systems.}
    \end{subfigure}
    \hfill
    \begin{subfigure}[b]{0.44\textwidth}
        \centering
        \includegraphics[width=\linewidth]{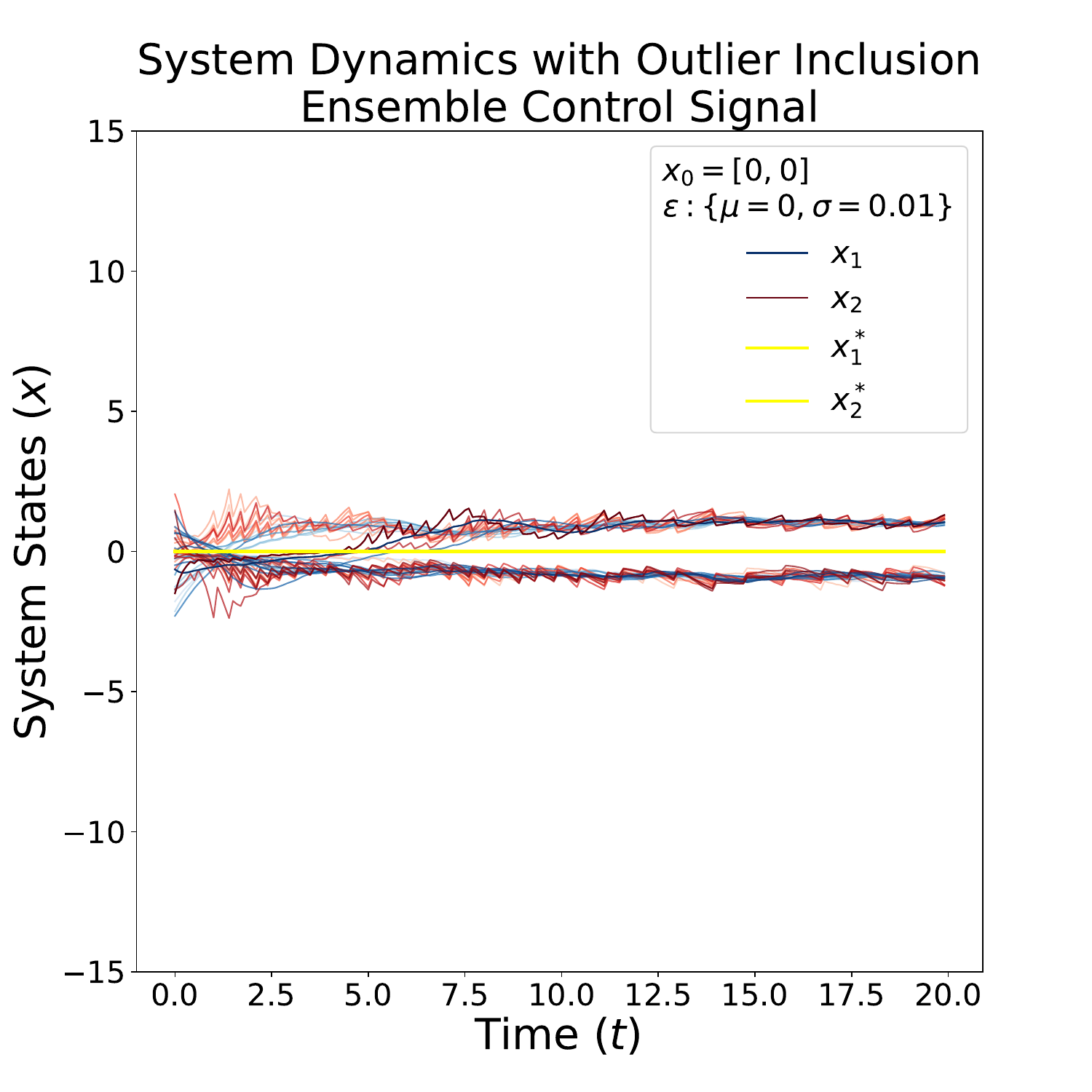}
        \caption{Validation of the system dynamics, with initial conditions $x_1$, $x_2$ = 0.0, resulting from applying a mean control signal ($\bar u$) that includes outlying systems.}
    \end{subfigure}\\
    \vspace{1em}
    \begin{subfigure}[b]{0.44\textwidth}
        \centering
        \includegraphics[width=\linewidth]{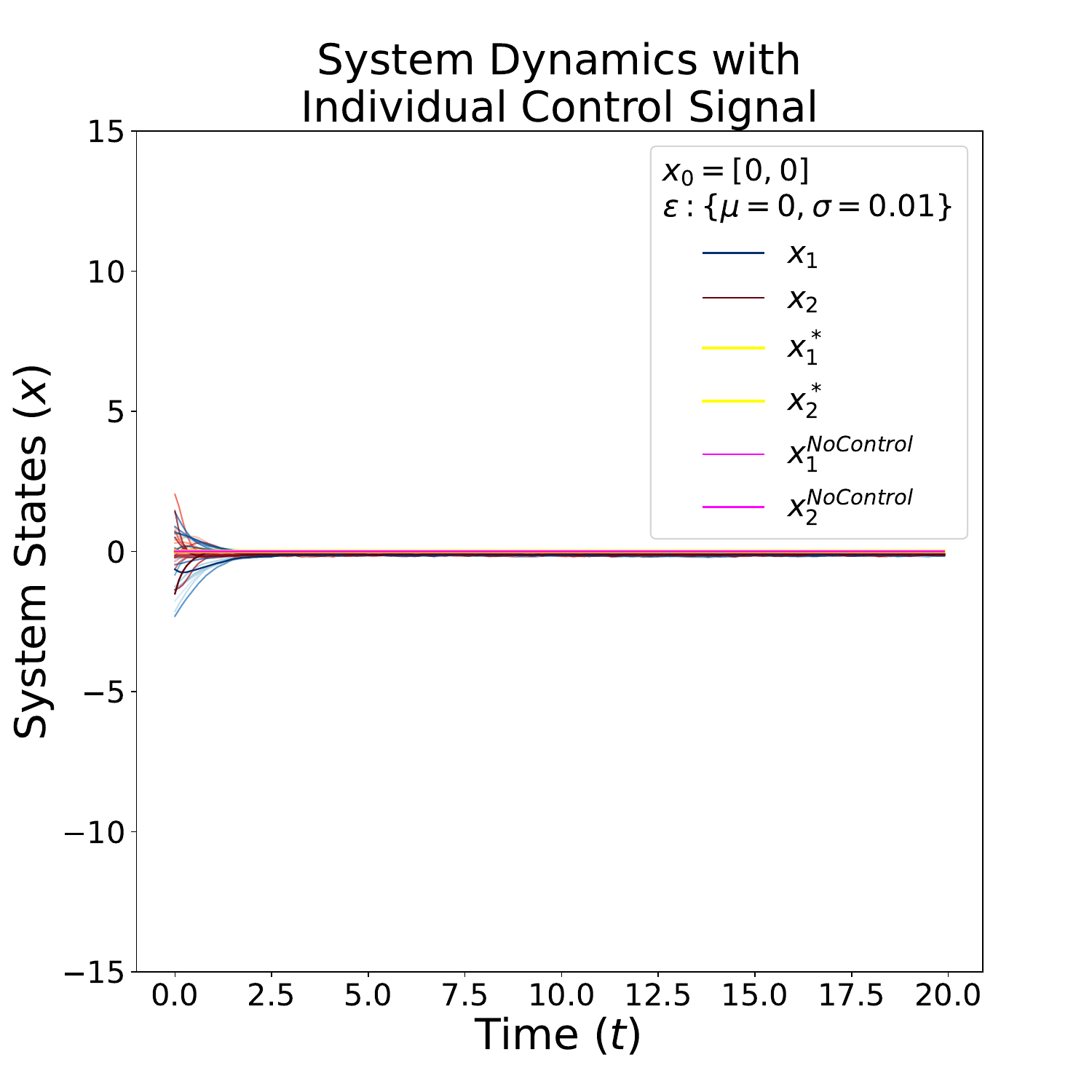}
        \caption{Validation of the system dynamics, with initial conditions $x_1$, $x_2$ = 0.0, resulting from applying each individual model's control signal ($\hat u^e$) to their respective system.}
    \end{subfigure}
\end{figure}

\begin{figure}[H]
    \ContinuedFloat
    \centering
    \begin{subfigure}[b]{0.44\textwidth}
        \centering
        \includegraphics[width=\linewidth]{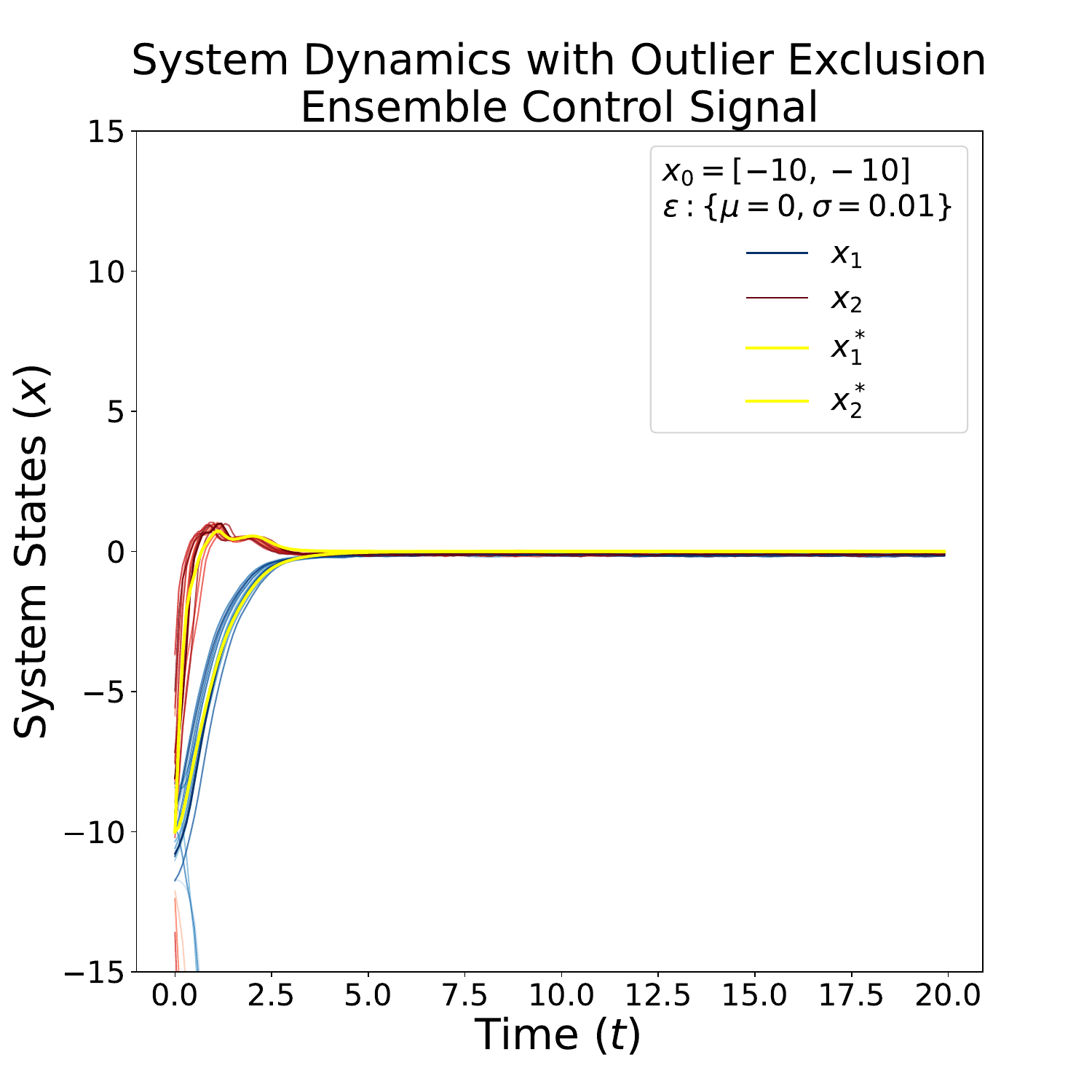}
        \caption{Validation of the system dynamics, with initial conditions $x_1$, $x_2$ = -10.0, resulting from applying a mean control signal ($\bar u$) that excludes outlying systems.}
    \end{subfigure}
    \hfill
    \begin{subfigure}[b]{0.44\textwidth}
        \centering
        \includegraphics[width=\linewidth]{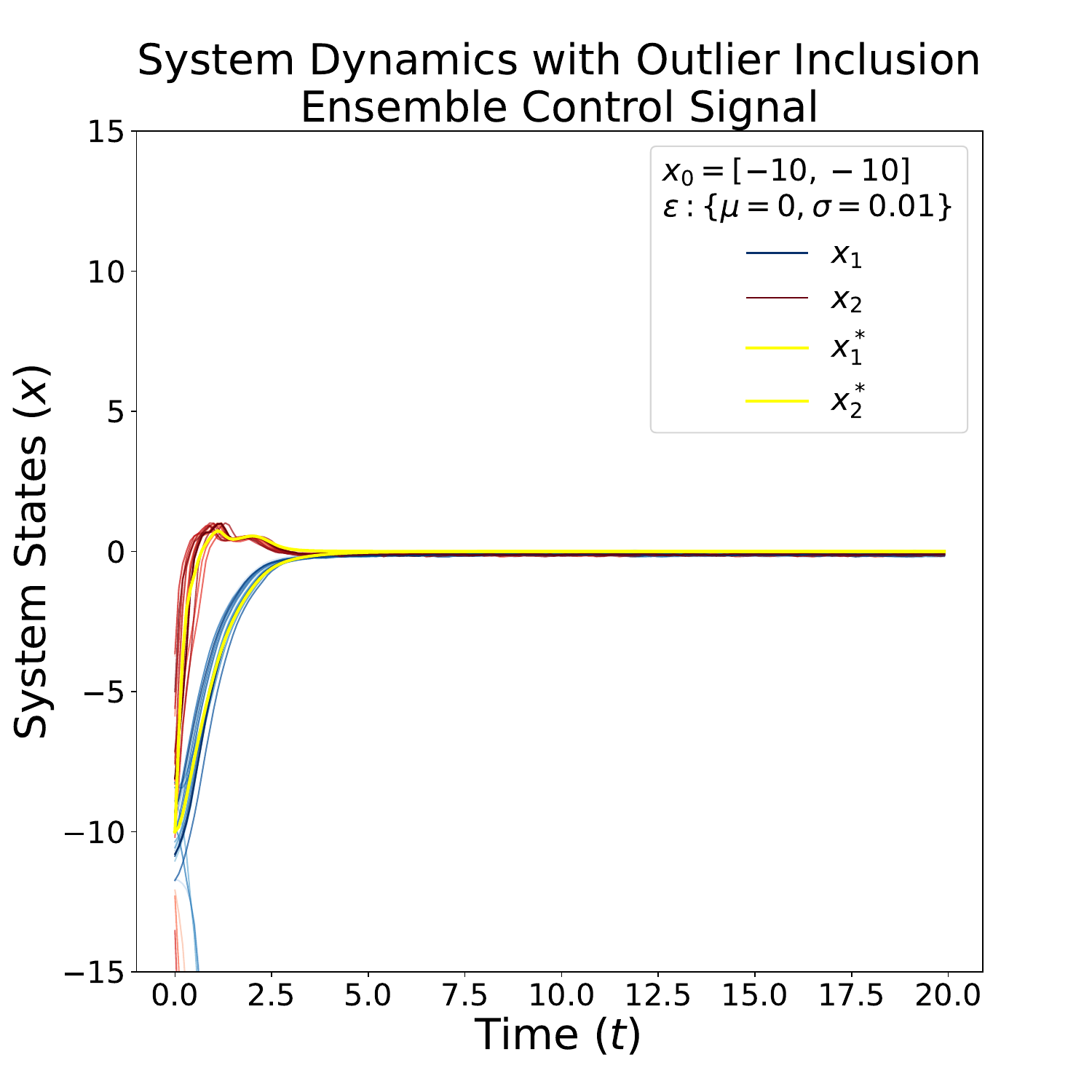}
        \caption{Validation of the system dynamics, with initial conditions $x_1$, $x_2$ = -10.0, resulting from applying a mean control signal ($\bar u$) that includes outlying systems.}
    \end{subfigure}\\
    \vspace{1em}
    \begin{subfigure}[b]{0.44\textwidth}
        \centering
        \includegraphics[width=\textwidth]{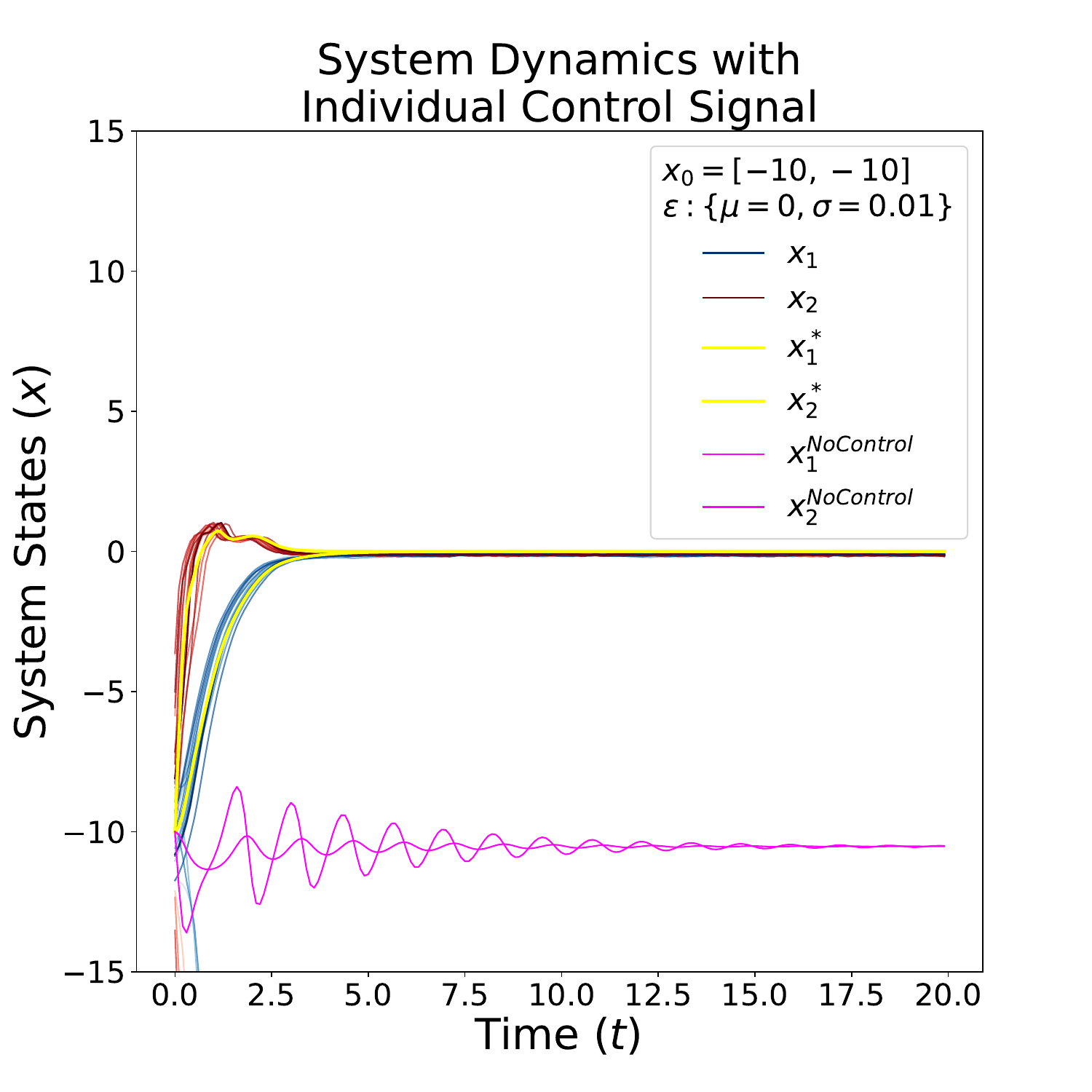}
        \caption{Validation of the system dynamics, with initial conditions $x_1$, $x_2$ = -10.0, resulting from applying each individual model's control signal ($\hat u^e$) to their respective system.}
    \end{subfigure}
\end{figure}

\bibliographystyle{elsarticle-num}
\bibliography{main}

\begin{thebibliography}{10}
\expandafter\ifx\csname url\endcsname\relax
  \def\url#1{\texttt{#1}}\fi
\expandafter\ifx\csname urlprefix\endcsname\relax\def\urlprefix{URL }\fi
\expandafter\ifx\csname href\endcsname\relax
  \def\href#1#2{#2} \def\path#1{#1}\fi

\bibitem{kirk2004optimal}
D.~E. Kirk, Optimal control theory: an introduction, Courier Corporation, 2004.

\bibitem{brunton_kutz_2019}
S.~L. Brunton, J.~N. Kutz, Data-Driven Science and Engineering: Machine Learning, Dynamical Systems, and Control, Cambridge University Press, 2019.
\newblock \href {https://doi.org/10.1017/9781108380690} {\path{doi:10.1017/9781108380690}}.

\bibitem{brunton2022nonlinearcontrol}
S.~L. Brunton, \href{https://app.dimensions.ai/details/publication/pub.1156592870}{Nonlinear control: Hamilton jacobi bellman (hjb) and dynamic programming}, CassyniHttps://cassyni.com/events/L6Vtg3maHUeTVjiXfqwTZf (2022).
\newblock \href {https://doi.org/10.52843/cassyni.4t5069} {\path{doi:10.52843/cassyni.4t5069}}.
\newline\urlprefix\url{https://app.dimensions.ai/details/publication/pub.1156592870}

\bibitem{khalil2002nonlinear}
H.~K. Khalil, Nonlinear systems third edition, Patience Hall 115 (2002).

\bibitem{Geering_2007}
H.~P. Geering, Optimal control with engineering applications, Springer, Berlin; New York, 2007.

\bibitem{hao2022physics}
Z.~Hao, S.~Liu, Y.~Zhang, C.~Ying, Y.~Feng, H.~Su, J.~Zhu, Physics-informed machine learning: A survey on problems, methods and applications, arXiv preprint arXiv:2211.08064 (2022).

\bibitem{raissi2019physics}
M.~Raissi, P.~Perdikaris, G.~E. Karniadakis, Physics-informed neural networks: A deep learning framework for solving forward and inverse problems involving nonlinear partial differential equations, Journal of Computational physics 378 (2019) 686--707.

\bibitem{dierks2009optimal}
T.~Dierks, S.~Jagannthan, Optimal control of affine nonlinear discrete-time systems, in: 2009 17th Mediterranean Conference on Control and Automation, IEEE, 2009, pp. 1390--1395.

\bibitem{dierks2010optimal}
T.~Dierks, S.~Jagannathan, Optimal control of affine nonlinear continuous-time systems, in: Proceedings of the 2010 American control conference, IEEE, 2010, pp. 1568--1573.

\bibitem{zargarzadeh2014adaptive}
H.~Zargarzadeh, T.~Dierks, S.~Jagannathan, Adaptive neural network-based optimal control of nonlinear continuous-time systems in strict-feedback form, International Journal of Adaptive Control and Signal Processing 28~(3-5) (2014) 305--324.

\bibitem{furfaro2022physics}
R.~Furfaro, A.~D'Ambrosio, E.~Schiassi, A.~Scorsoglio, Physics-informed neural networks for closed-loop guidance and control in aerospace systems, in: AIAA SCITECH 2022 Forum, 2022, p. 0361.

\bibitem{sutton2018reinforcement}
R.~S. Sutton, A.~G. Barto, Reinforcement Learning: An Introduction, MIT Press, 2018.

\bibitem{lillicrap2015continuous}
T.~P. Lillicrap, et~al., Continuous control with deep reinforcement learning, arXiv preprint arXiv:1509.02971 (2015).

\bibitem{silver2016mastering}
D.~Silver, et~al., Mastering the game of go with deep neural networks and tree search, Nature 529~(7587) (2016) 484--489.

\bibitem{chua2018deep}
K.~Chua, et~al., Deep reinforcement learning in a handful of trials using probabilistic dynamics models, arXiv preprint arXiv:1805.12114 (2018).

\bibitem{willard2022integrating}
J.~Willard, et~al., Integrating scientific knowledge with machine learning for engineering and environmental systems, Nature Reviews Earth \& Environment 3~(9) (2022) 513--532.

\bibitem{raj2025deepoperator}
A.~Raj, S.~Bun, K.~Srinivasa, C.~E. Gudumotou, A.~Sarshar, \href{https://www.sciencedirect.com/science/article/pii/S0010465525003558}{Deep operator networks for bayesian parameter estimation in pdes}, Computer Physics Communications 317 (2025) 109853.
\newblock \href {https://doi.org/https://doi.org/10.1016/j.cpc.2025.109853} {\path{doi:https://doi.org/10.1016/j.cpc.2025.109853}}.
\newline\urlprefix\url{https://www.sciencedirect.com/science/article/pii/S0010465525003558}

\bibitem{Barry-Straume2022May}
J.~Barry-Straume, A.~Sarshar, A.~A. Popov, A.~Sandu, Physics-informed neural networks for pde-constrained optimization and control, Communications on Applied Mathematics and Computation (2025).
\newblock \href {https://doi.org/10.1007/s42967-025-00499-x} {\path{doi:10.1007/s42967-025-00499-x}}.

\bibitem{karniadakis2021physics}
G.~E. Karniadakis, et~al., Physics-informed machine learning, Nature Reviews Physics 3~(6) (2021) 422--440.

\bibitem{dietterich2000ensemble}
T.~G. Dietterich, Ensemble methods in machine learning, International Workshop on Multiple Classifier Systems (2000) 1--15.

\bibitem{agrachev2022control}
A.~Agrachev, Control systems on manifolds and problems of ensemble controllability, Proceedings of the Steklov Institute of Mathematics 316~(1) (2022) 16--41.

\bibitem{Staritsyn2022}
M.~Staritsyn, N.~Pogodaev, R.~Chertovskih, F.~L. Pereira, \href{https://doi.org/10.1109%2Flcsys.2021.3089139}{Feedback maximum principle for ensemble control of local continuity equations: An application to supervised machine learning}, {IEEE} Control Systems Letters 6 (2022) 1046--1051.
\newblock \href {https://doi.org/10.1109/lcsys.2021.3089139} {\path{doi:10.1109/lcsys.2021.3089139}}.
\newline\urlprefix\url{https://doi.org/10.1109%2Flcsys.2021.3089139}

\bibitem{ChilledWater2021}
S.~Deng, Z.~Chen, F.~Kuang, C.~Yang, W.~Gui, Optimal control of chilled water system with ensemble learning and cloud edge terminal implementation, IEEE Transactions on Industrial Informatics 17~(11) (2021) 7839--7848.
\newblock \href {https://doi.org/10.1109/TII.2021.3057943} {\path{doi:10.1109/TII.2021.3057943}}.

\bibitem{lee2019ensemble}
K.~Lee, Z.~Wang, B.~Vlahov, H.~Brar, E.~A. Theodorou, Ensemble bayesian decision making with redundant deep perceptual control policies, in: 2019 18th IEEE International Conference On Machine Learning And Applications (ICMLA), IEEE, 2019, pp. 831--837.

\bibitem{agrachev2016ensemble}
A.~Agrachev, T.~Chambrion, Ensemble controllability of nonlinear control systems, ESAIM: Control, Optimisation and Calculus of Variations 22~(2) (2016) 356--367.

\bibitem{Beauchard_2010}
K.~Beauchard, J.-M. Coron, P.~Rouchon, \href{https://doi.org/10.1007\%2Fs00220-010-1008-9}{Controllability issues for continuous-spectrum systems and ensemble controllability of bloch equations}, Communications in Mathematical Physics 296~(2) (2010) 525--557.
\newblock \href {https://doi.org/10.1007/s00220-010-1008-9} {\path{doi:10.1007/s00220-010-1008-9}}.
\newline\urlprefix\url{https://doi.org/10.1007\%2Fs00220-010-1008-9}

\bibitem{fotiadis2023physics}
F.~Fotiadis, K.~G. Vamvoudakis, A physics-informed neural networks framework to solve the infinite-horizon optimal control problem, in: 2023 62nd IEEE Conference on Decision and Control (CDC), IEEE, 2023, pp. 6014--6019.

\bibitem{schiassi2022bellman}
E.~Schiassi, A.~D'Ambrosio, R.~Furfaro, Bellman neural networks for the class of optimal control problems with integral quadratic cost, IEEE Transactions on Artificial Intelligence 5~(3) (2022) 1016--1025.

\bibitem{dey2024data}
S.~Dey, H.~Xu, A data-enabled dual learning based online receding horizon safe-critical control for nonlinear systems under uncertainty, in: NAECON 2024-IEEE National Aerospace and Electronics Conference, IEEE, 2024, pp. 310--315.

\bibitem{nishimura2024combined}
K.~Nishimura, H.~Hoshino, E.~Furutani, Combined plant and control co-design via solutions of hamilton-jacobi-bellman equation based on physics-informed learning, arXiv preprint arXiv:2409.02345 (2024).

\bibitem{mukherjee2023actor}
A.~Mukherjee, J.~Liu, Actor-critic methods using physics-informed neural networks: Control of a 1d pde model for fluid-cooled battery packs, arXiv preprint arXiv:2305.10952 (2023).

\bibitem{bryson2018applied}
A.~E. Bryson, Applied optimal control: optimization, estimation and control, Routledge, 2018.

\bibitem{sontag2013mathematical}
E.~D. Sontag, Mathematical control theory: deterministic finite dimensional systems, Vol.~6, Springer Science \& Business Media, 2013.

\bibitem{evans2022partial}
L.~C. Evans, Partial differential equations, Vol.~19, American Mathematical Society, 2022.

\bibitem{bellman1957dynamic}
R.~Bellman, R.~Corporation, K.~M.~R. Collection, \href{https://books.google.com/books?id=wdtoPwAACAAJ}{Dynamic Programming}, Rand Corporation research study, Princeton University Press, 1957.
\newline\urlprefix\url{https://books.google.com/books?id=wdtoPwAACAAJ}

\bibitem{bertsekas1996dynamic}
D.~P. Bertsekas, Dynamic programming and optimal control, Journal of the Operational Research Society 47~(6) (1996) 833--833.

\bibitem{pontryagin2018mathematical}
L.~S. Pontryagin, Mathematical theory of optimal processes, Routledge, 2018.

\bibitem{rackauckas2017differentialequations}
C.~Rackauckas, Q.~Nie, Differentialequations.jl--a performant and feature-rich ecosystem for solving differential equations in julia, Journal of Open Research Software 5~(1) (2017) 15.

\bibitem{tsitouras2011runge}
C.~Tsitouras, Runge--kutta pairs of order 5 (4) satisfying only the first column simplifying assumption, Computers \& Mathematics with Applications 62~(2) (2011) 770--775.

\bibitem{bhattacharjee2024improving}
A.~Bhattacharjee, A.~A. Popov, A.~Sarshar, A.~Sandu, Improving adam through an implicit-explicit (imex) time-stepping approach, Journal of Machine Learning for Modeling and Computing 5~(3) (2024).
\newblock \href {https://doi.org/10.1615/jmachlearnmodelcomput.2024053508} {\path{doi:10.1615/jmachlearnmodelcomput.2024053508}}.

\bibitem{towers2024gymnasium}
M.~Towers, A.~Kwiatkowski, J.~Terry, J.~U. Balis, G.~De~Cola, T.~Deleu, M.~Goul{\~a}o, A.~Kallinteris, M.~Krimmel, A.~KG, et~al., Gymnasium: A standard interface for reinforcement learning environments, arXiv preprint arXiv:2407.17032 (2024).

\end{thebibliography}

\end{document}